\documentclass{article}
\usepackage[utf8]{inputenc}

\usepackage{times}
\usepackage{graphicx}
\usepackage{amsmath}
\usepackage{amssymb}
\usepackage{algorithm}
\usepackage{algorithmic}
\usepackage{subcaption}

\usepackage{hyperref}
\usepackage{multirow}

\usepackage{enumitem}

\DeclareMathOperator*{\minimise}{minimise}
\DeclareMathOperator*{\subject}{subject}
\DeclareMathOperator*{\sgn}{sgn}
\DeclareMathOperator*{\argmin}{argmin}
\DeclareMathOperator*{\tr}{tr}

\begin{document}

\title{Side Information in Robust Principal Component Analysis: Algorithms and Applications}

\author{Niannan Xue\\
{\tt\small n.xue15@imperial.ac.uk}
\and
Yannis Panagakis\\
{\tt\small i.panagakis@imperial.ac.uk}
\and
Stefanos Zafeiriou\\
{\tt\small s.zafeiriou@imperial.ac.uk }
}

\date{}

\maketitle

\begin{abstract}
Robust Principal Component Analysis (RPCA) aims at recovering a low-rank subspace from grossly corrupted high-dimensional (often visual) data and is a cornerstone in many machine learning and computer vision applications. Even though RPCA has been shown to be very successful in solving many rank minimisation problems, there are still cases where degenerate or suboptimal solutions are obtained. This is likely to be remedied by taking into account of domain-dependent prior knowledge. In this paper, we propose two models for the RPCA problem with the aid of side information on the low-rank structure of the data. The versatility of the proposed methods is demonstrated by applying them to four applications, namely background subtraction, facial image denoising, face and facial expression recognition. Experimental results on synthetic and five real world datasets indicate the robustness and effectiveness of the proposed methods on these application domains, largely outperforming six previous approaches.
\end{abstract}

\section{Introduction}

Principal Component Pursuit (PCP) as proposed in \cite{Candes11,Chandrasekaran11} and its variants e.g. \cite{Aravkin14,Shang14,Xu12,Zhou10,Bao12,Cabral2013} are the current methods of choice for recovering a low-rank subspace from a set of grossly corrupted and possibly incomplete high-dimensional data. PCP employs the nuclear norm and the $l_1$ norm (convex surrogates of the rank and sparsity constraints, respectively) in order to approximate the original $l_0$ norm regularised rank minimisation problem. In particular, under certain conditions (such as the restricted isometry property \cite{Candes08}), the relaxation gap is zero and rank minimisation is equivalent to nuclear norm minimisation. However, these conditions rarely hold for real-world visual data and PCP thus occasionally yields degenerate or suboptimal solutions. To alleviate this, it is advantageous for PCP to take into account of domain-dependent prior knowledge \cite{Jiao15}, i.e. side information \cite{Ziv76}.
 
The use of side information has been studied in the context of matrix completion \cite{Chiang15,Xu13} and compressed sensing \cite{Mota14}. Recently, side information has been applied to the PCP framework in the \textit{noiseless} case \cite{Sagonas14,KaiYang16}. In particular, an error-free orthogonal column space was used to drive a PCP-based deformable image alignment algorithm \cite{Sagonas14}. More generally, \cite{KaiYang16} used both a column and a row space as side information and the algorithm had to recover the weights of their interaction. The main limitation of such methods is that they require a set of clean, noise-free data samples in order to determine the column and/or row spaces of the low-rank component. Clearly, these data are are difficult to find in practice.
 
In this paper, we investigate the idea of using a \textit{noisy} approximation of the low-rank component to guide PCP. Knowledge regarding the low-rank component, albeit noisy, is available in many applications. In background subtraction, we may find some frames of the video that do not contain changes and therefore may be used to accurately estimate the background. Another example concerns the problem of disentangling identity and expression components in expressive faces, where the low-rank component is roughly similar to the neutral face. Note that side information which has the same form as the source is already subject to wide-spread usage. Watermark detection methods require a reference image to identify ownership \cite{Cox97}. Automated photo tagging explores visually similar social images \cite{Wu09}. Locality preserving projection can be enhanced by exploiting similar pairs of patterns \cite{An08}. Spatial and temporal correlation can improve signal recovery algorithms in compressive imaging \cite{Stankovic09}. In content-based image retrieval, historical feedback log data can help retrieve semantically relevant images \cite{Zhang12}. Low-resolution images can help adapt a high-resolution compressive sensing system \cite{Warnell15}. Near-accurate fingerprint or DNA can be used as side information to hack a biometric authentication system \cite{Kang15}.

Our contributions are summarised as follows:
\begin{itemize}
    \item A novel convex program is proposed to use side information, which is a noisy approximation of the low-rank component, within the PCP framework with a provably convergent solver.
    \item Furthermore, we extend our proposed PCP model using side information to exploit prior knowledge regarding the column and row spaces of the low-rank component in a more general algorithmic framework.
    \item We demonstrate the applicability and effectiveness of the proposed approaches in several applications, namely background subtraction, facial image denoising as well as face recognition and facial expression classification.
    \item We also show that our proposed methods can mitigate the transductive constraint of RPCA. With side information, training can be performed on fewer samples and hence reducing the computational cost.\\
\end{itemize}
\textbf{\textit{Notations}} Lowercase letters denote scalars and uppercase letters denote matrices, unless otherwise stated. For norms of matrix $\mathbf{A}$, $\Vert \mathbf{A}\Vert_F$ is the Frobenius norm; $\Vert \mathbf{A}\Vert_*$ is the nuclear norm; and $\Vert \mathbf{A}\Vert_\infty$ is the maximum absolute value among all matrix entries. Moreover, $\langle \mathbf{A},\mathbf{B}\rangle$ represents tr($\mathbf{A}^T\mathbf{B}$) for real matrices $\mathbf{A},\mathbf{B}$. Additionally, $\sigma_i$ is the $i$th largest singular value of a matrix and $\sigma_{j\%}$ is the singular value at the $j$th percentile.

\section{Related work} 

The problem of incorporating side information in estimating low-rank components can be stated as follows. Suppose that there is a matrix $\mathbf{L}_0\in\mathbb{R}^{n_1\times n_2}$ with rank $r\ll$ min($n_1,n_2$) and a sparse matrix $\mathbf{S}_0\in\mathbb{R}^{n_1\times n_2}$ with entries of arbitrary magnitude. If we are provided with the data matrix
\begin{equation}
    \mathbf{M} = \mathbf{L}_0+\mathbf{S}_0,
\end{equation}
and additional side information, how can we recover the low-rank component $\mathbf{L}_0$ and the sparse noise $\mathbf{S}_0$ accurately by taking advantage of the side information?

One the first methods for incorporating side information was proposed in the context of deformable face alignment \cite{Sagonas14}. The RAPS algorithm assumes that we have available an orthogonal column space $\mathbf{X}\in\mathbb{R}^{n_1\times d_1}$, where $d_1\le n_1$, and
\begin{equation}
    \begin{split}
        &\minimise_{\mathbf{G},\mathbf{S}}\quad\Vert \mathbf{G}\Vert_*+\lambda\Vert\mathbf{S}\Vert_1\\
        &\subject\text{ to}\ \ \,\mathbf{X}\mathbf{G}+\mathbf{S}=\mathbf{M}.\\
    \end{split}
\end{equation}

A generalisation of the above was proposed as Principal Component Pursuit with Features (PCPF) in \cite{KaiYang16} where further row spaces $\mathbf{Y}\in\mathbb{R}^{n_2\times d_2}$ were assumed to be available with $d_2\le n_2$, and 
\begin{equation}
    \begin{split}
        &\minimise_{\mathbf{H},\mathbf{S}}\quad\Vert \mathbf{H}\Vert_*+\lambda\Vert\mathbf{S}\Vert_1\\
        &\subject\text{ to}\ \ \,\mathbf{X}\mathbf{H}\mathbf{Y}^T+\mathbf{S}=\mathbf{M}.\\
    \end{split}
\end{equation}

\cite{Shahdi15,Shahdi16} incorporate structural knowledge into RPCA by adding spectral graph regularisation. Given the graph Laplacian $\mathbf{\Phi}$ of each data similarity graph, Robust PCA on Graphs (RPCAG) and Fast Robust PCA on Graphs (FRPCAG) add an additional tr$(\mathbf{L}\mathbf{\Phi}\mathbf{L}^T)$ term to the PCP objective for the low-rank component $\mathbf{L}$. The main drawback of the above mentioned models is that the side information needs to be accurate and noiseless, which is not trivial in practical scenarios.

\section{Robust Principal Component Analysis Using Side Information}

In this section, the proposed RPCA models with side information are introduced. In particular, we propose to incorporate the side information into PCP by using the trace distance of the difference between the low-rank component and the noisy estimate, which is reasonable if their difference is of low rank. However, we show empirically (Section \ref{experiment}) that it also works if the difference is full-rank. This may be attributed to the fact that the trace distance is a natural distance metric between two dissimilar distributions from Kolmogorov$-$Smirnov statistics \cite{Nielsen10}. Besides that, this is a generalisation of the compressed sensing with side information where the $l_1$ norm has been used in order to measure the distance of the target signal with prior information \cite{Mota14}.

\subsection{The PCPS model}

Assuming that a noisy estimate of the low-rank component of the data $\mathbf{W}\in\mathbb{R}^{n_1\times n_2}$ is available, we propose the following model of PCP using side information (PCPS):
\begin{equation}
    \begin{split}
        &\minimise_{\mathbf{L},\mathbf{S}}\quad\Vert \mathbf{L}\Vert_*+\kappa\Vert \mathbf{L}-\mathbf{W}\Vert_*+\lambda\Vert\mathbf{S}\Vert_1\\
        &\subject\text{ to}\ \ \,\mathbf{L}+\mathbf{S}=\mathbf{M},\\
    \end{split}
\end{equation}
where $\kappa>0,\lambda>0$ are parameters that weigh the effects of side information and noise sparsity. 

The proposed PCPS can be revamped to generalise the previous attempt of PCPF by the following objective of PCPS with features (PCPSF):
\begin{equation}\label{eq:PCPSF}
    \begin{split}
        &\minimise_{\mathbf{H},\mathbf{S}}\quad\Vert\mathbf{H}\Vert_*+\kappa\Vert\mathbf{H}-\mathbf{D}\Vert_*+\lambda\Vert\mathbf{S}\Vert_1\\
        &\subject\text{ to}\ \ \,\mathbf{X}\mathbf{H}\mathbf{Y}^T+\mathbf{S}=\mathbf{M},\quad \mathbf{X}\mathbf{D}\mathbf{Y}^T=\mathbf{W},\\
    \end{split}
\end{equation}
where $\mathbf{H}\in\mathbb{R}^{d_1\times d_2},\mathbf{D}\in\mathbb{R}^{d_1\times d_2}$ are bilinear mappings for the recovered low-rank matrix $\mathbf{L}$ and side information $\mathbf{W}$ respectively. Note that the low-rank matrix $\mathbf{L}$ is recovered from the optimal solution ($\mathbf{H}^*,\mathbf{S}^*$) to objective (\ref{eq:PCPSF}) via $\mathbf{L}=\mathbf{X}\mathbf{H}^*\mathbf{Y}^T$. If side information $\mathbf{W}$ is not available, PCPSF reduces to PCPF by setting $\kappa$ to zero. If the features $\mathbf{X},\mathbf{Y}$ are not present either, PCP can be restored by fixing both of them at identity. However, when only the side information $\mathbf{W}$ is accessible, objective (\ref{eq:PCPSF}) is transformed back into PCPS.

\subsection{The algorithm}

If we substitute $\mathbf{E}$ for $\mathbf{H}-\mathbf{D}$ and orthogonalise $\mathbf{X}$ and $\mathbf{Y}$, the optimisation problem (\ref{eq:PCPSF}) is identical to the following convex but non-smooth problem:
\begin{equation}\label{eq:PCPSF2}
    \begin{split}
        &\minimise_{\mathbf{H},\mathbf{S}}\quad\Vert\mathbf{H}\Vert_*+\kappa\Vert\mathbf{E}\Vert_*+\lambda\Vert\mathbf{S}\Vert_1\\
        &\subject\text{ to}\ \ \,\mathbf{X}\mathbf{H}\mathbf{Y}^T+\mathbf{S}=\mathbf{M},\quad\mathbf{E}-\mathbf{H} = -\mathbf{X}^T\mathbf{W}\mathbf{Y},\raisetag{1 cm}\\
    \end{split}
\end{equation}
which is amenable to the multi-block alternating direction method of multipliers (ADMM).

The corresponding augmented Lagrangian of (\ref{eq:PCPSF2}) is:
\begin{equation}\label{eq:PCPSFl}
\begin{split}
&l(\mathbf{H},\mathbf{E},\mathbf{S},\mathbf{Z},\mathbf{N})=\Vert\mathbf{H}\Vert_*+\kappa\Vert\mathbf{E}\Vert_*+\lambda\Vert\mathbf{S}\Vert_1\\&+\langle\mathbf{Z},\mathbf{M}-\mathbf{S}-\mathbf{X}\mathbf{H}\mathbf{Y}^T\rangle +\frac{\mu}{2}\Vert\mathbf{M}-\mathbf{S}-\mathbf{X}\mathbf{H}\mathbf{Y}^T\Vert_F^2\\&+\langle\mathbf{N},\mathbf{H}-\mathbf{E}-\mathbf{X}^T\mathbf{W}\mathbf{Y}\rangle+\frac{\mu}{2}\Vert\mathbf{H}-\mathbf{E}-\mathbf{X}^T\mathbf{W}\mathbf{Y}\Vert_F^2,\raisetag{1.8cm}\\
\end{split}
\end{equation}
where $\mathbf{Z}\in\mathbb{R}^{n_1\times n_2}$ and $\mathbf{N}\in\mathbb{R}^{d_1\times d_2}$ are Lagrange multipliers and $\mu$ is the learning rate.

The ADMM operates by carrying out repeated cycles of updates till convergence. During each cycle, $\mathbf{H},\mathbf{E},\mathbf{S}$ are updated serially by minimising (\ref{eq:PCPSFl}) with other variables fixed. Afterwards, Lagrange multipliers $\mathbf{Z},\mathbf{N}$ are updated at the end of each iteration. Direct solutions to the single variable minimisation subproblems rely on  the shrinkage and the singular value thresholding operators \cite{Candes11}. Let $\mathcal{S}_\tau(a)\equiv\sgn(a)\max(|a|-\tau,0)$ serve as the shrinkage operator, which naturally extends to matrices,  $\mathcal{S}_\tau(\mathbf{A})$, by applying it to matrix $\mathbf{A}$ element-wise. Similarly, let $\mathcal{D}_\tau(\mathbf{A})\equiv \mathbf{U}\mathcal{S}_\tau(\mathbf{\Sigma})\mathbf{V}^T$ be the singular value thresholding operator on real matrix $\mathbf{A}$, with $\mathbf{A}=\mathbf{U}\mathbf{\Sigma}\mathbf{V}^T$ being the singular value decomposition (SVD) of $\mathbf{A}$.

Minimising (\ref{eq:PCPSFl}) w.r.t. $\mathbf{H}$ at fixed $\mathbf{E},\mathbf{S},\mathbf{Z},\mathbf{N}$ is equivalent to the following:
\begin{equation}
    \arg\min_{\mathbf{H}}\ \Vert\mathbf{H}\Vert_*+\mu\Vert\mathbf{P}-\mathbf{X}\mathbf{H}\mathbf{Y}^T\Vert_F^2,
\end{equation}
where $\mathbf{P} = \frac{1}{2}(\mathbf{M}-\mathbf{S}+\mathbf{W}+\frac{1}{\mu}\mathbf{Z}+\mathbf{X}(\mathbf{E}-\frac{1}{\mu}\mathbf{N})\mathbf{Y}^T)$. Its solution is shown to be $\mathbf{X}^T\mathcal{D}_{\frac{1}{2\mu}}(\mathbf{P})Y$. Furthermore, for $\mathbf{E}$,
\begin{equation}
    \arg\min_{\mathbf{E}}\ l=\arg\min_{\mathbf{E}}\ \kappa\Vert\mathbf{E}\Vert_*+\frac{\mu}{2}\Vert\mathbf{Q}-\mathbf{E}\Vert_F^2,
\end{equation}
where $\mathbf{Q}=\mathbf{H}-\mathbf{X}^T\mathbf{W}\mathbf{Y}+\frac{1}{\mu}N$, whose update rule is $\mathcal{D}_{\frac{\kappa}{\mu}}(\mathbf{Q})$, and for $\mathbf{S}$,
\begin{equation}
    \arg\min_{\mathbf{S}}\ l=\arg\min_{\mathbf{S}}\ \lambda\Vert\mathbf{S}\Vert_1+\frac{\mu}{2}\Vert\mathbf{R}-\mathbf{S}\Vert_F^2,
\end{equation}
where $\mathbf{R}=\mathbf{M}-\mathbf{X}\mathbf{H}\mathbf{Y}^T+\frac{1}{\mu}\mathbf{Z}$ with a closed-form solution $\mathcal{S}_{\lambda\mu^{-1}}(\mathbf{R})$. Finally, Lagrange multipliers are updated as usual:
\begin{equation}
    \mathbf{Z}= \mathbf{Z}+\mu(\mathbf{M}-\mathbf{S}-\mathbf{X}\mathbf{H}\mathbf{Y}^T),
\end{equation}
\begin{equation}
    \mathbf{N} =\mathbf{N} +\mu (\mathbf{H}-\mathbf{E}-\mathbf{X}^T\mathbf{W}\mathbf{Y}).
\end{equation}
The overall algorithm is summarised in Algorithm \ref{alg:PCPSFa}.
\begin{algorithm}[h]
   \caption{ADMM solver for PCPSF}
   \label{alg:PCPSFa}
\begin{algorithmic}[1]
   \REQUIRE Observation $\mathbf{M}$, side information $\mathbf{W}$, features $\mathbf{X},\mathbf{Y}$, parameters $\kappa,\lambda>0$, scaling ratio $\alpha>1$.
   \STATE {\bfseries Initialize:} $\mathbf{Z}=0$, $\mathbf{N}=\mathbf{E}=\mathbf{H}=0$, $\mu=\frac{1}{\Vert\mathbf{M}\Vert_2}$.
   \WHILE{not converged}
   \STATE $\mathbf{S}=\mathcal{S}_{\lambda\mu^{-1}}(\mathbf{M}-\mathbf{X}\mathbf{H}\mathbf{Y}^T+\frac{1}{\mu}\mathbf{Z})$
   \STATE $\mathbf{H}=\mathbf{X}^T\mathcal{D}_{\frac{1}{2\mu}}(\frac{1}{2}(\mathbf{M}-\mathbf{S}+\mathbf{W}+\frac{1}{\mu}\mathbf{Z}+\mathbf{X}(\mathbf{E}-\frac{1}{\mu}\mathbf{N})\mathbf{Y}^T))\mathbf{Y}$
   \STATE $\mathbf{E}=\mathcal{D}_{\kappa\mu^{-1}}(\mathbf{H}-\mathbf{X}^T\mathbf{W}\mathbf{Y}+\frac{1}{\mu}\mathbf{N})$
   \STATE $\mathbf{Z} = \mathbf{Z}+\mu(\mathbf{M}-\mathbf{S}-\mathbf{X}\mathbf{H}\mathbf{Y}^T)$
   \STATE $\mathbf{N} =\mathbf{N} +\mu (\mathbf{H}-\mathbf{E}-\mathbf{X}^T\mathbf{W}\mathbf{Y})$
   \STATE $\mu = \mu\times\alpha$
   \ENDWHILE
   \ENSURE $\mathbf{L}=\mathbf{X}\mathbf{H}\mathbf{Y}^T$, $\mathbf{S}$
\end{algorithmic}
\end{algorithm}

\subsection{Complexity and convergence}

Orthogonalisation of the features $\mathbf{X},\mathbf{Y}$ via the Gram-Schmidt process has an operation count of $O(n_1d_1^2)$ and $O(n_2d_2^2)$ respectively. The $\mathbf{H}$ update in Step $4$ is the most costly step of each iteration in Algorithm \ref{alg:PCPSFa}. Specifically, the SVD required in the singular value thresholding action dominates with $O(\min(n_1n_2^2,n_1^2n_2))$ complexity.

It has been recently established that for a 3-block separable convex minimisation problem, the direct extension of the ADMM achieves global convergence with linear convergence rate if one block in the objective is sub-strongly monotonic \cite{Sun16}. In our case, it can be shown that $\Vert\mathbf{S}\Vert_1$ processes such sub-strong monotonicity. We have also used the fast continuation technique to increase $\mu$ incrementally for accelerated superlinear performance. The cold start initialisation strategies for variables $\mathbf{H},\mathbf{E}$ and Lagrange multipliers $\mathbf{Z},\mathbf{N}$ are described in \cite{Boyd11}. Besides, we have scheduled $\mathbf{S}$ to be updated first. As for stopping criteria, we have employed the Karush-Kuhn-Tucker (KKT) feasibility conditions. Namely, within a maximum number of $1000$ iterations, when the maximum of $\Vert\mathbf{M}-\mathbf{S}_k-\mathbf{X}\mathbf{H}_k\mathbf{Y}^T\Vert_F/\Vert\mathbf{M}\Vert_F$ and $\Vert\mathbf{H}_k -\mathbf{E}_k - \mathbf{X}^T\mathbf{W}\mathbf{Y}\Vert_F/\Vert\mathbf{M}\Vert_F$ dwindles from a pre-defined threshold $\epsilon$, the algorithm is terminated, where $k$ signifies values at the $k$\textsuperscript{th} iteration.

\section{Experimental results}
\label{experiment}

In this section, we illustrate the enhancement made by side information through both numerical simulations and real-world applications. First, we compare the recoverability of our proposed algorithms with state-of-the-art methods for incorporating features or dictionaries, i.e. PCPF \cite{KaiYang16} and RAPS \cite{Sagonas14} on synthetic data as well as the baseline PCP \cite{Candes11} when there are no features available. Second, we show how powerful side information can be for the task of object segmentation in video pre-processing. Third, we demonstrate that side information is instructive in the low-dimensionality face modeling from images of different illuminations. Last, we reveal that the more accurately reconstructed expressions in the light of side information lead to better emotion classification.

For RAPS, clean subspace $\mathbf{X}$ is used instead of the observation $\mathbf{M}$ itself as the dictionary in LRR \cite{Liu13}. PCP is solved via the inexact ALM and the heuristics for predicting the dimension of principal singular space is not adopted here due to its lack of validity on uncharted real data. We also include Partial Sum of Singular Values (PSSV) \cite{Oh16} in our comparison for its stated advantage in view of the limited number of expression observations available.

\subsection{Parameter calibration}

The process of tuning the algorithmic parameters for various models is described in the supplementary material. Although theoretical determination of $\kappa$ and $\lambda$ is beyond the scope of this paper, we nevertheless provide empirical guidance based on extensive experiments. $\lambda=1/\sqrt{\max(n_1,n_2)}$ for a general matrix of dimension $n_1\times n_2$ from PCP works well for both of our proposed models. $\kappa$ depends on the quality of the side information. When the side information is accurate, a large $\kappa$ should be selected to capitalise upon the side information as much as possible, whereas when the side information is improper, a small $\kappa$ should be picked to sidestep the dissonance caused by the side information. Here, we have discovered that a $\kappa$ value of $0.2$ works best with synthetic data and a value of $0.5$ is suited for public video sequences. It is worth emphasising again that prior knowledge of the structural information about the data yields more appropriate values for $\kappa$ and $\lambda$.

\subsection{Phase transition on synthetic datasets}

\begin{figure*}[h]
	\begin{center}
		\includegraphics[width=1\linewidth]{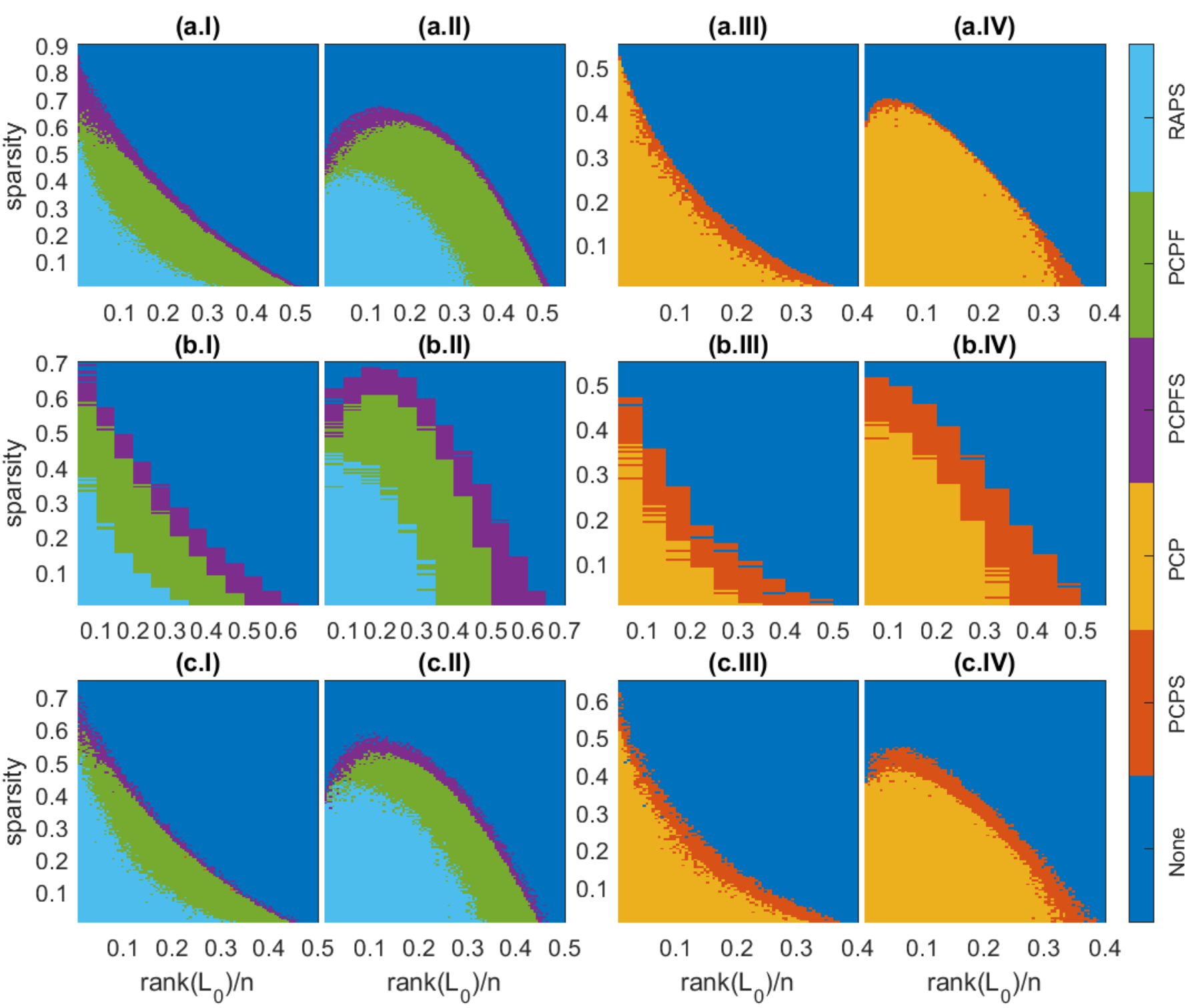}
	\end{center}
	\caption{Domains of recovery by various algorithms: \textbf{(I,III)} for random signs and \textbf{(II,IV)} for coherent signs. \textbf{(a)} for entry-wise corruptions, \textbf{(b)} for deficient ranks and \textbf{(c)} for distorted singular values.}
	\label{fig:simulation}
\end{figure*}

We now focus on the recoverability problem, i.e. recovering matrices of varying ranks from errors of varying sparsity. True low-rank matrices are created via $\mathbf{L}_0=\mathbf{J}\mathbf{K}^T$, where $200\times r$ matrices $\mathbf{J},\mathbf{Y}$ have independent elements drawn randomly from a Gaussian distribution of mean $0$ and variance $5\cdot10^{-3}$ so $r$ is the rank of $\mathbf{L}_0$. Next, we generate $200\times200$ error matrices $\mathbf{S}_0$, which possess $\rho_s\cdot200^2$ non-zero elements located randomly within the matrix. We consider two types of entries for $\mathbf{S}_0$: Bernoulli $\pm1$ and $\mathcal{P}_\Omega(\sgn(\mathbf{L}_0))$, where $\mathcal{P}$ is the projection operator and $\Omega$ is the support set of $\mathbf{S}_0$. $\mathbf{M}=\mathbf{L}_0+\mathbf{S}_0$ thus becomes the simulated observation. For each $(r,\rho_s)$ pair, three observations are constructed. The recovery is successful if for all these three problems, 
\begin{equation}
    \frac{\Vert\mathbf{L}-\mathbf{L}_0\Vert_F}{\Vert\mathbf{L}_0\Vert_F}<10^{-3}
\end{equation}
from the recovered $\mathbf{L}$. In addition, let $\mathbf{L}_0=\mathbf{U}\mathbf{\Sigma}\mathbf{V}^T$ be the SVD of $\mathbf{L}_0$. Feature $\mathbf{X}$ is formed by randomly interweaving column vectors of $\mathbf{U}$ with $d$ arbitrary orthonormal bases for the null space of $\mathbf{U}^T$, while permuting the expanded columns of $\mathbf{V}$ with $d$ random orthonormal bases for the kernel of $\mathbf{V}^T$ forms feature $\mathbf{Y}$. Hence, the feasibility conditions are fulfilled: $\mathbb{C}(\mathbf{X})\supseteq\mathbb{C}(\mathbf{L}_0)$, $\mathbb{C}(\mathbf{Y})\supseteq\mathbb{C}(\mathbf{L}_0^T)$, where $\mathbb{C}$ is the column space operator.

\textbf{Entry-wise corruptions.} For these trials, we construct the side information by directly adding small Gaussian noise to each element of $\mathbf{L}_0$: $l_{ij}\rightarrow l_{ij}+\mathcal{N}(0,2.5r\cdot10^{-9})$, $i,j=1,2,\cdots,200$. As a result, the standard deviation of the error in each element is $1\%$ of that among the elements themselves. On average, the Frobenius percent error, $\Vert\mathbf{W}-\mathbf{L}_0\Vert_F/\Vert\mathbf{L}_0\Vert_F$, is $1\%$. Such side information is genuine in regard to the fact that classical PCA with accurate rank is not able to eliminate the noise \cite{Shabalin13}. For $d=10$, Figures \ref{fig:simulation}(a.I) and \ref{fig:simulation}(a.II) plot results from PCPF, RAPS and PCPSF. On the other hand, the situation with no available features is investigated in Figures \ref{fig:simulation}(a.III) and \ref{fig:simulation}(a.IV) for PCP and PCPS. The frontier of PCPF has been advanced by PCPSF everywhere for both sign types. Especially at low ranks, errors with much higher density can be removed. Without features, PCPS surpasses PCP by and large with significant expansion at small sparsity for both cases.

\textbf{Deficient ranks.} Now we first make a new matrix $\mathbf{\Sigma}'$ by retaining only the singular values from $\sigma_1$ to $\sigma_{90\%}$ in $\mathbf{\Sigma}$. Then, side information is constructed according to $\mathbf{W}=\mathbf{U}\mathbf{\Sigma}'\mathbf{V}^T$, aka hard thresholding. As rank increases, Frobenius percent error of $\mathbf{W}$ decreases from $23.3\%$ to $5.8\%$ sublinearly. Figures \ref{fig:simulation}(b.I) and \ref{fig:simulation}(b.II) show results from PCPF, RAPS and PCPSF where $d$ is again kept at $10$. The corresponding cases with no features are presented in Figures \ref{fig:simulation}(b.III) and \ref{fig:simulation}(b.IV) for PCP and PCPS. Notwithstanding the most spurious side information, PCPSF and PCPS have reclaimed the largest region unattainable by PCPF and PCP respectively for the two signs.

\textbf{Distorted singular values.} Here, we produce the matrix $\mathbf{\Sigma}'$ by adding Gaussian noise to singular values in $\mathbf{\Sigma}$: $\sigma_i\rightarrow\sigma_i+0.01\cdot\mathcal{N}(0,\sigma_i^2)$ for all $i$. Next, side information is formed by $\mathbf{W}=\mathbf{U}\mathbf{\Sigma}'\mathbf{V}^T$. The mean Frobenius percent error in $\mathbf{W}$ is $1\%$. With $d$ relaxed to $50$, recoverability diagrams for PCPF, RAPS, PCPSF and PCP, PCPS are drawn in Figures (c.I), (c.II) and (c.III), (c.IV). We observe substantial growth of recoverability for PCPS over PCP across the full range of ranks. And with features, there is still omniscient gain in recoverablity for PCPSF against PCPF, which is marked at low ranks.

We remark that in unrecoverable areas, PCPS and PCPSF still obtain much smaller values of $\Vert\mathbf{L}-\mathbf{L}_0\Vert_F$. In view of the marginal improvement of RAPS contrasted with PCPF and PCPSF, we will not consider it any longer. Results from RPCAG and PSSV are worse than PCP (see the supplementary material). FRPCAG fails to recover anything at all.

\subsection{Face denoising under variable illumination}

\begin{figure*}[h]
	\begin{center}
		\includegraphics[width=1\linewidth]{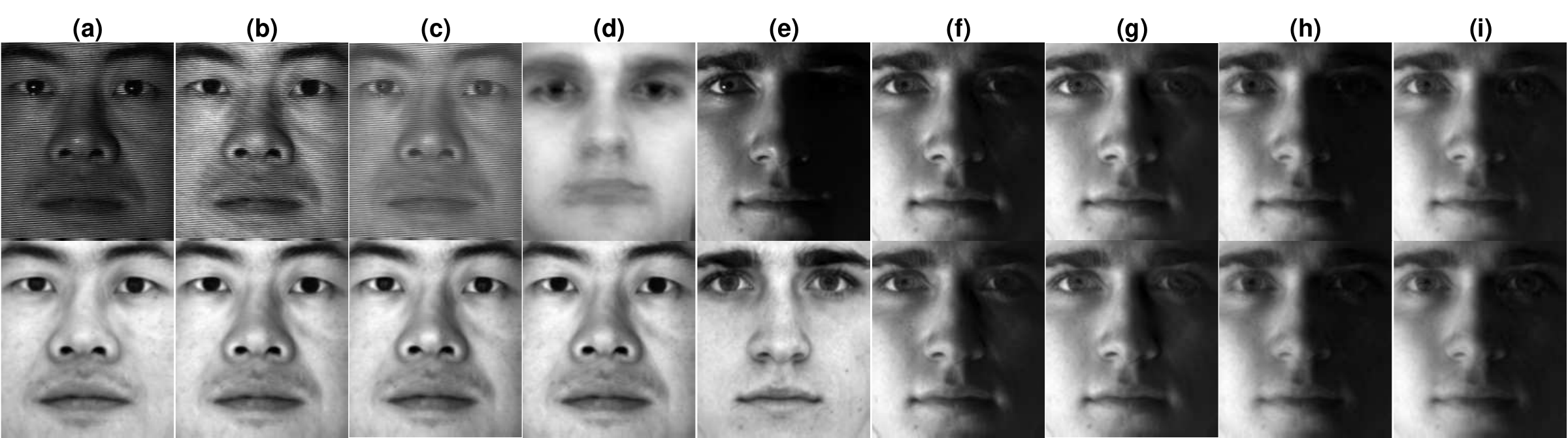}
	\end{center}
	\caption{Comparison of face denoising ability: In row I, \textbf{(a, e)} sample frames from subjects 2 and 33; \textbf{(b, f)} single-person PCP; \textbf{(c, g)} single-person PCPF; \textbf{(h, i)} multi-person PCP and PCPF; \textbf{(d)} average of other subjects. In row II, \textbf{(a, e)} average of a single subject; \textbf{(b, f)} single-person PCPS; \textbf{(c, g)} single-person PCPSF; \textbf{(h, i)} multi-person PCPS and PCPSF; \textbf{(d)} PCPS using the side information above.}
	\label{fig:Yale}
\end{figure*}

It has been previously proved that a convex Lambertian surface under distant and isotropic lighting has an underlying model that spans a 9-D linear subspace. Albeit faces can be described as Lambertian, it is only approximate and harmonic planes are not real images due to negative pixels. In addition, theoretical lighting conditions cannot be realised and there are unavoidable occlusion and albedo variations. It is thus more natural to decompose facial image formation as a low-rank component for face description and a sparse component for defects. What is more, we suggest that further boost to the performance of facial characterisation can be gained by leveraging an image which faithfully represents the subject.

We consider images of a fixed pose under different illuminations from the extended Yale B database for testing. Ten subjects were randomly chosen and all 64 images were studied for each person. For single-person experiments, $32556\times64$ observation matrices were formed by vectorising each $168\times192$ image and the side information was chosen to be the average of all images, tiled to the same size as the observation matrix for each subject. For the multiperson experiment, both single-person observation and side information matrices were concatenated into $32556\times640$ matrices respectively.

For PCPF and PCPSF to run, we learn the feature dictionary following an approach by Vishal et al.~\cite{Patel12}. In a nutshell, the feature learning process can be treated as a sparse encoding problem. More specifically, we simultaneously seek a dictionary $\mathbf{D}\in\mathbb{R}^{n_1\times c}$ and a sparse representation $\mathbf{B}\in\mathbb{R}^{c\times n_2}$ such that:
\begin{equation}
    \begin{split}
        &\minimise_{\mathbf{D},\mathbf{B}}\quad\Vert\mathbf{M}-\mathbf{D}\mathbf{B}\Vert_F^2\\
        &\subject\text{ to}\ \ \,\gamma_i\le t\text{  for }i=1\dots n_2,\\
    \end{split}
\end{equation}
where $c$ is the number of atoms, $\gamma_i$'s count the number of non-zero elements in each sparsity code and $t$ is the sparsity constraint factor. This can be solved by the K-SVD algorithm. Here, feature $\mathbf{X}$ is the dictionary $\mathbf{D}$, feature $\mathbf{Y}$ corresponds to a similar solution using the transpose of the observation matrix as input and the sparse codes are irrelevant. For implementation details, we set $c$ to $40$, $t$ to $40$ and used $10$ iterations. Because K-SVD could not converge in reasonable time for the multiperson experiment, we resorted to classical PCA applied to the observation matrix to obtain features $\mathbf{X},\mathbf{Y}$ of dimension $400$.

As a visual illustration, two challenging cases are exhibited in Figure \ref{fig:Yale} (PSSV, RPCAG, FRPCAG do not improve upon PCP and are shown in the supplementary material). For subject 2, it is clearly evident that PCPS and PCPSF outperform the best existing methods through the complete elimination of acquisition faults. More surprisingly, PCPSF even manages to restore the flash in the pupils that is not present in the side information. For subject 33, PCPS indubitably reconstructs a more vivid left eye than that from PCP which is only discernible. With that said, PCPSF still prevails by uncovering more shadows, especially around the medial canthus of the left eye, and revealing a more distinct crease in the upper eyelid as well a more translucent iris. We also notice that results from the single-person experiment outdo their counterparts from the multiperson experiment. Thence, we will focus on a single subject alone. 

\begin{figure}[h]
	\begin{center}
		\includegraphics[width=.8\linewidth]{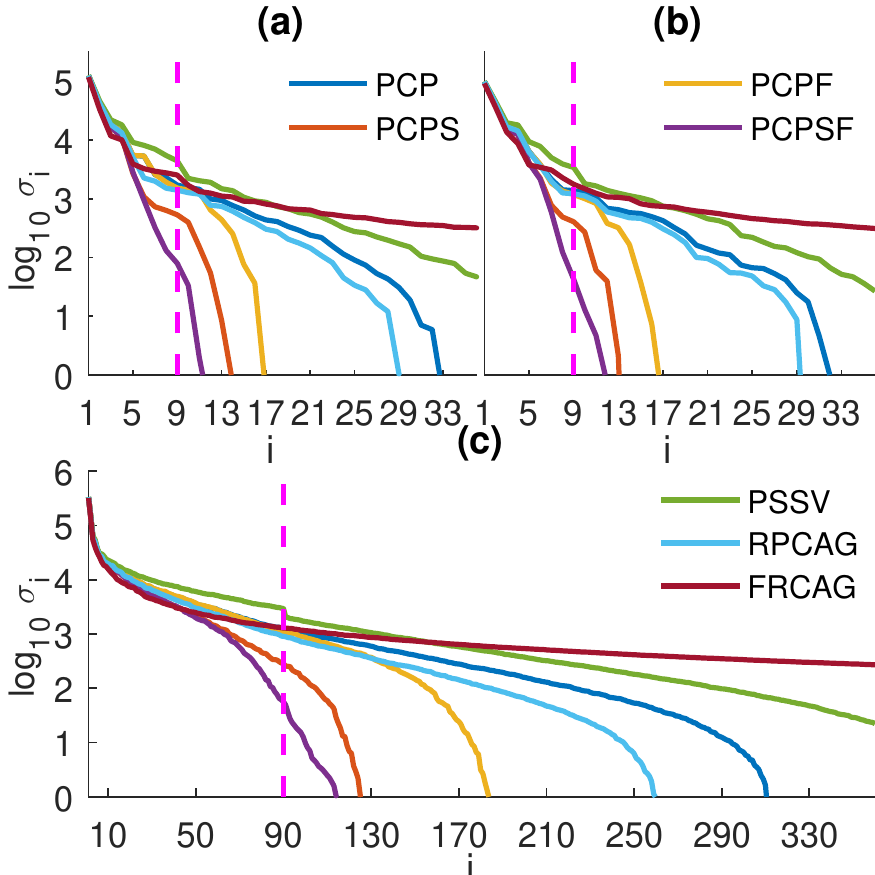}
	\end{center}
	\caption{Log-scale singular values of the denoised matrices: \textbf{(a)} subject 2; \textbf{(b)} subject 33; \textbf{(c)} all subjects.}
	\label{fig:rank}
\end{figure}

To quantitatively verify the improvement made by our proposed approaches, we examine the structural information contained within the deionised eigenfaces. Singular values of the recovered low-rank matrices from all algorithms are plotted in Figure \ref{fig:rank}. Singular values decease most sharply for PCPSF followed by PCPS. By the theoretical limit, they are orders of magnitude smaller than those values from other methods. This validates our proposed approaches.

We further unmask the strength of PCPS by considering the stringent side information made of the average of other subjects. Surprisingly, PCPS still manages to remove the noise recovering an authentic image (see Figure \ref{fig:Yale} \textbf{(d)}).

\subsection{Background subtraction from surveillance video}

\begin{figure*}[h]
\begin{center}
\includegraphics[width=1\linewidth]{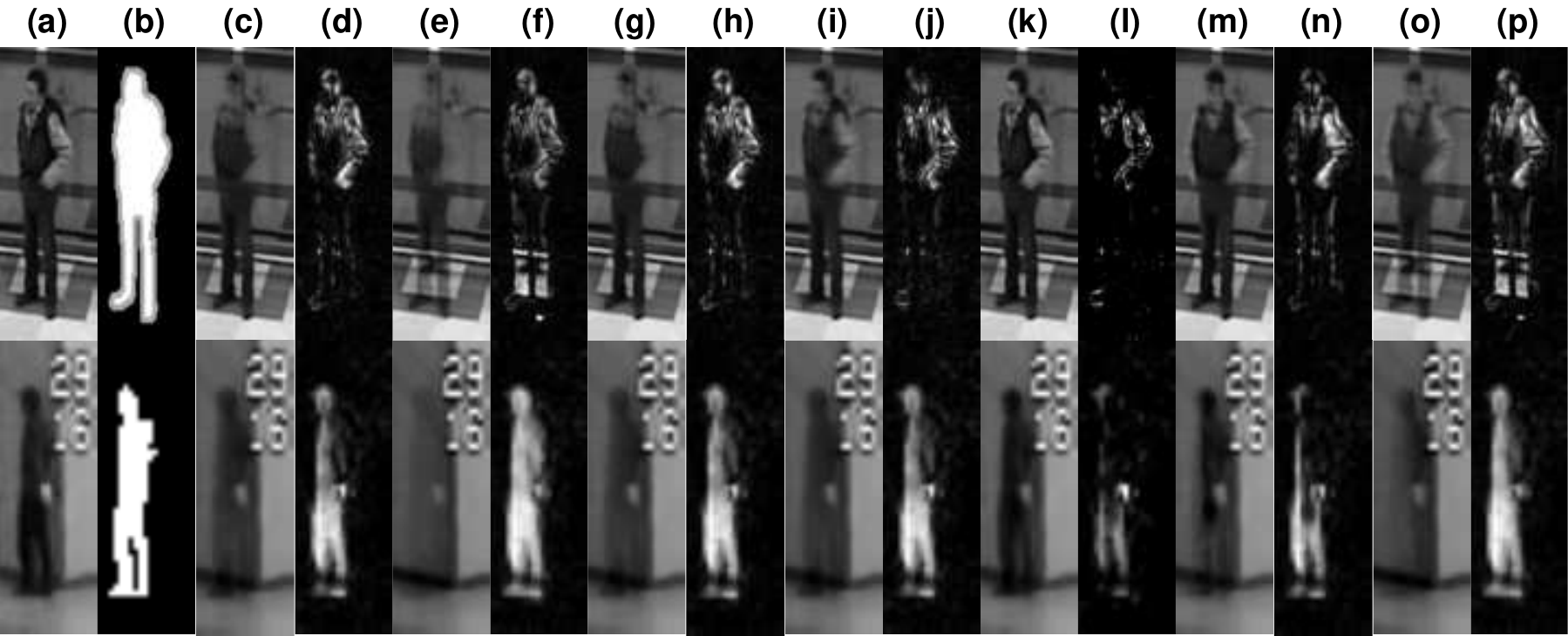}
\end{center}
   \caption{Background subtraction results for two sample frames, PETS in row I and Airport in row II: \textbf{(a)} original images; \textbf{(b)} ground truth;
   \textbf{(c,d)} PCP; \textbf{(e,f)} PCPS; \textbf{(g,h)} PSSV; \textbf{(i,j)} RPCAG; \textbf{(k,l )} FRPCAG; \textbf{(m,n)} PCP (60 frames); \textbf{(o,p)} PCPS (60 frames).}
    \label{fig:bs}
\end{figure*}

In automated video analytics, object detection is instrumental in object tracking, activity recognition and behaviour understanding. Practical applications include surveillance, traffic control, robotic operation, etc., where foreground objects can be people, vehicles, products and so forth. Background subtraction segments moving objects by calculating the pixel-wise difference between each video frame and the background. For a static camera, the background is almost static, while the foreground objects are mostly moving. Consequently, a decomposition into a low-rank component for the background and a sparse component for foreground objects is a valid model for such dynamics. Indeed, if the only change in the background is illumination, then the matrix representation of vectorised backgrounds has a rank of $1$. It has been demonstrated that PCP is quite effective for such a low-rank matrix analysis problem \cite{Candes11}. Nevertheless, through the application of our proposed algorithm to such a background-foreground separation scenario, we show that useful side information can help achieve better background restoration.

One video sequence from the PETS 2006 dataset and one from the I2R dataset were utilised for evaluation. Each consists of scenes at a hall where people walk intermittently. 200 consecutive frames of $720\times576$ resolution grayscale images were stacked by columns into a $414720\times200$ observation matrix from the first video and 200 frames of $176\times144$ images from the second video were stacked into another $25344\times200$ observation matrix. Two side information arrays comprised columns that are copies of a vectorised photo which contains an empty hallway. To commence object detection, PCP and PCPS were first run to extract the backgrounds. Then objects were recovered by calculating the absolute values of the difference between the original frame and the estimated background. Since parameters for dictionary learning need exhaustive search, we will not be comparing PCPF and PCPSF for what follows. 

\begin{figure}[h]
	\begin{center}
		\includegraphics[width=.8\linewidth]{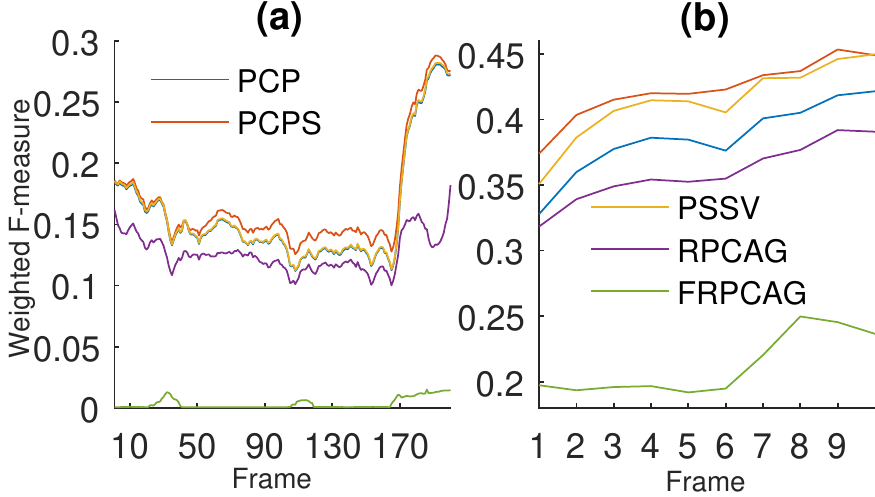}
	\end{center}
	\caption{Weighted F-measure scores: \textbf{(a)} PETS; \textbf{(b)} Airport.}
	\label{fig:measure}
\end{figure}

We quantitatively compare the performance of the competing methods according to the weighted F-measure \cite{Margolin14} against manually annotated bounding boxes provided as the ground truth. The resulting scores for each frame are presented in Figure \ref{fig:measure}. From the consistently higher precision statistics, the merit of PCPS over PCP is confirmed. For qualitative reference, representative images of the recovered background and foreground from all methods are listed in Figure \ref{fig:bs} (For space reasons, we have only included the most noticeable sector. See the supplementary material for whole images.). PCP and its variants only partially detect infrequent moving objects, people who stop moving for extended periods of time, leaving ghost artifacts in the background. In contrast, PCPS segments a fairly sharp silhouette of slowly moving objects to produce a much cleaner background, promoting its novelty.

To further unravel of the robustness of our propositions, shortened videos from PETS and Airport consisting of 60 frames are analysed via PCPS. Figures \ref{fig:bs} \textbf{(c,d)} \& \textbf{(o,p)} show that PCPS with less input can achieve comparative or better results than PCP with more input. This suggests that the transductive constraint of RPCA no longer applies because with the help of side information we can run PCPS on fewer frames rather than the entire collection every time new observation arrives. The speed-ups for PETS and Airport are $2.44\times$ and $2.62\times$ respectively.

\subsection{Face and facial expression recognition}

\begin{figure}[h]
	\begin{center}
		\includegraphics[width=.65\linewidth]{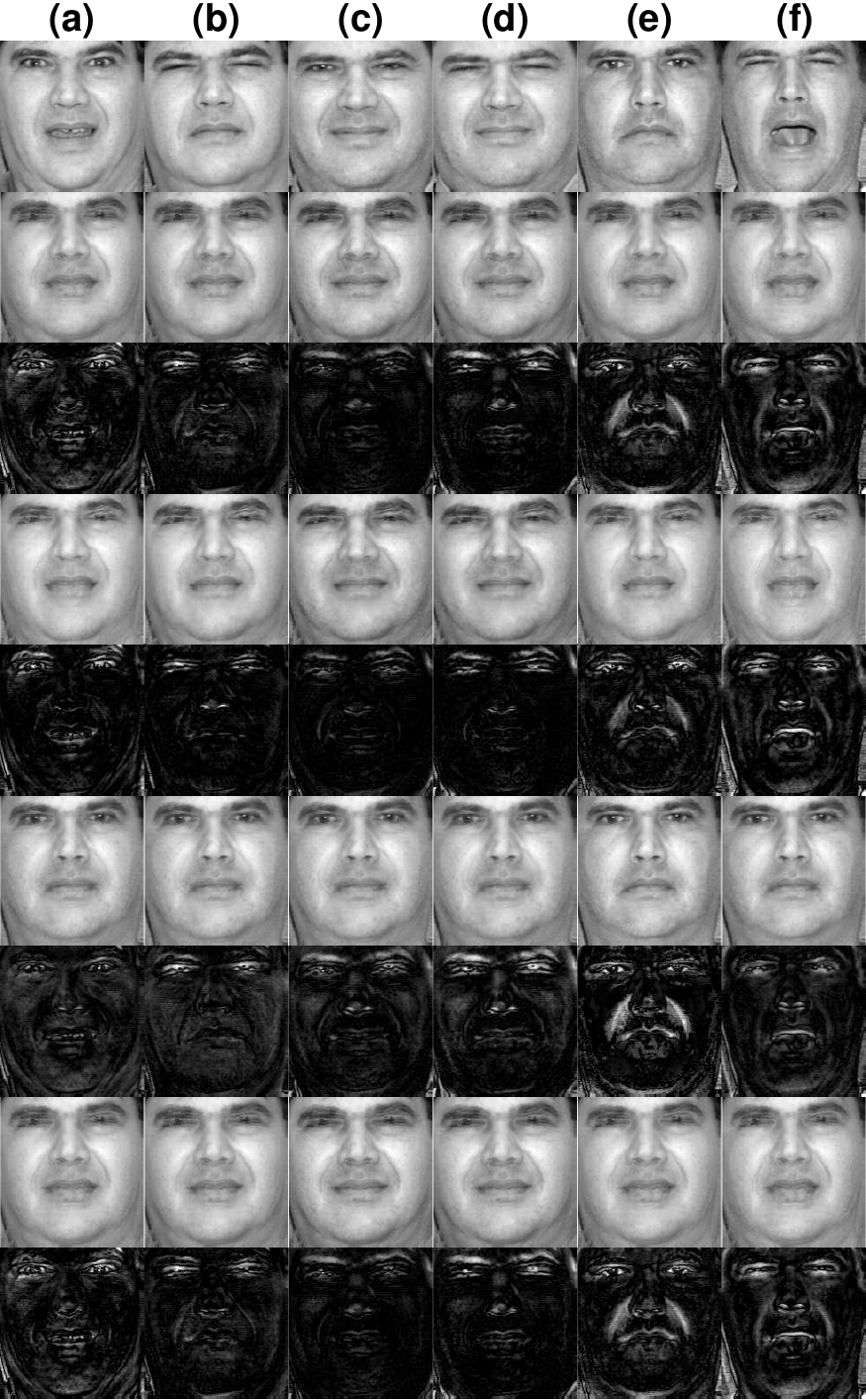}
	\end{center}
	\caption{Expression extraction for a single subject: Expressive faces reside in row I. Identity classes produced by PCP, PSSV, PCPS, RPCAG are in rows II, IV, VI, VIII. The complementary expression components are depicted in rows III, V, VII, IX.}
	\label{fig:express}
\end{figure}

Recent research has established that an expressive face can be treated as a neutral face plus a sparse expression component \cite{Taheri13}, which is identity-independent due to its constituent local non-rigid motions, i.e. action units. This is central to computer vision as it enables human emotion classification from such visual cues. We will demonstrate how the accurate reconstruction of facial expressions guided by side information ameliorates classification analysis.

To begin with, evaluation was effected on the CMU Multi-PIE dataset. Aligned and cropped $165\times172$ images of frontal pose and normal lighting from 54 subjects were used. We batch-processed each subject forming a $28380\times6$ observation matrix to extract expressions: Neutral, Smile, Surprise, Disgust, Scream and Squint. For each subject, side information was offered by a sextet of neutral face repetitions. Archetypal expressions recovered by PCP, PCPS, PSSV, RPCAG are laid out in Figure \ref{fig:express} (the restricted number of expressions disallows FRPCAG). It is noteworthy that local appearance changes separated by PCPS are the most salient which paves the way for better classification. We avail ourselves of the multi-class RBF-kernel SVM and the SRC \cite{Wright09} to map expressions to emotions. 9-fold cross-validation results are reported in Table \ref{table:class}. PCPS leads PCP by a fair margin with PSSV, RPCAG underperforming PCP. 

\begin{table}[h]
\begin{center}
\small
\begin{tabular}{|c|c|c|c|c|}
\hline
Algorithm & PCP & PSSV & PCPS & RPCAG \\
\hline
Non-linear SVM & 78.40 & 74.69 & \textbf{79.94} & 77.16 \\
\hline
SRC & 79.01 & 74.38 & \textbf{82.72} & 79.01 \\
\hline
\end{tabular}
\caption{Classification accuracy (\%) on the Multi-PIE dataset for PCP, PSSV, PCPS and RPCAG by means of non-linear SVM and SRC learning.}
\label{table:class}
\end{center}
\vspace{-.2cm}
\end{table}

Lastly, the CK+ dataset was incorporated to assess the joint face and expression recognition capabilities of various algorithms. Each test image is sparsely coded via a dictionary of both identities and universal expressions (Anger, Disgust, Fear, Happiness, Sadness and Surprise). The least resulting reconstruction residual thereupon determines its identity or expression. We refer readers to \cite{Georgakis16} for the exact problem set-up and implementation details. Table \ref{table:joint} collects the computed recognition rates. Altough RPCAG and FRPCAG are superior than PCP as expected, PCPS performs distinctly better than all others.

\begin{table}[h]

\begin{center}
\small
\begin{tabular}{|c|c|c|c|c|c|}
\hline
Algorithm & PCP & PSSV & PCPS & RPCAG & FRPCAG\\
\hline
Identity & 87.35 & 87.05 & \textbf{95.23} & 89.77 & 90.98\\
\hline
Expression & 49.24 & 45.30 & \textbf{67.50} & 58.26 & 57.73\\
\hline
\end{tabular}
\caption{Recognition rates (\%) for joint identity \& expression recognition averaged over 10 trials on CK+}
\label{table:joint}
\end{center}
\vspace{-.2cm}
\end{table}

\section{Conclusion}

In this paper, we have, for the first time, assimilated side information of the same format as observation into the framework of Robust Principal Component Analysis based on trace norms. Existing extensions of subspace features have also been successfully amalgamated in a convex fashion. Extensive experiments have shown that our algorithms not only perform better where Robust PCA is effective but also remain potent when Robust PCA fails. Directions for future research include generalising to the tensor case and to components of multiple scales.

\clearpage

\appendix

\section{Parameter calibration}

In order to tune the algorithmic parameters, we first conduct a benchmark experiment as follows: a low-rank matrix $\mathbf{L}_0$ is generated from $\mathbf{L}_0=\mathbf{J}\mathbf{K}^T$, where $\mathbf{J},\mathbf{K}\in\mathbb{R}^{200\times10}$ have entries from a $\mathcal{N}(0,0.005)$ distribution; a $200\times200$ sparse matrix $\mathbf{S}_0$ is generated by randomly setting $38,000$ entries to zero with others taking values of $\pm1$ with equal probability. 

If $\mathbf{X}$ is set as the left-singular vectors of $\mathbf{L}_0$ and $\mathbf{Y}$ is set as the right-singular vectors of $\mathbf{L}_0$, then a scaling ratio $\alpha = 1.1$, a tolerance threshold $\epsilon = 10^{-7}$ and a maximum step size $\mu=10^{18}$ to avoid ill-conditioning can bring PCP, RAPS, PCPF to convergence with a recovered $\mathbf{L}$ of rank $10$, a recovered $\mathbf{S}$ of sparsity $5\%$ and an accuracy $\|\mathbf{L}-\mathbf{L}_0\|_F/\|\mathbf{L}_0\|_F$ on the order of $10^{-6}$. Hereafter, we will adopt these parameter settings for PCP, RAPS, PCPF and will apply them to PCPS and PCPSF as well. PSSV also uses these parameter settings as done similarly in \cite{Oh16}.

\begin{figure}[t]
	\vspace{-2.2cm}
	\begin{center}
		\includegraphics[width=.8\linewidth]{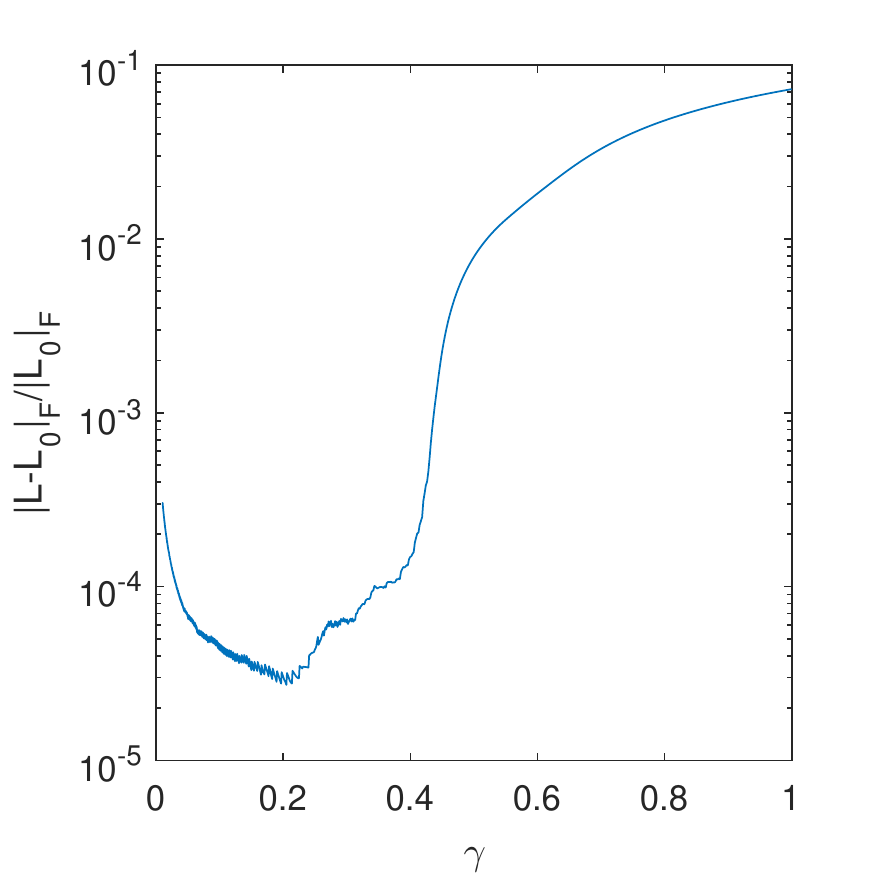}
	\end{center}
	\vspace{-.6cm}
	\caption{Relative error ($\frac{\|\mathbf{L}-\mathbf{L}_0\|_F}{\|\mathbf{L}_0\|_F}$) of RPCAG for $\gamma\in[0.01,1]$.}
	\label{fig:rpcagParam}
	\vspace{-.3cm}
\end{figure}

\begin{figure}[b]
	\vspace{-.3cm}
	\begin{center}
		\includegraphics[width=.8\linewidth]{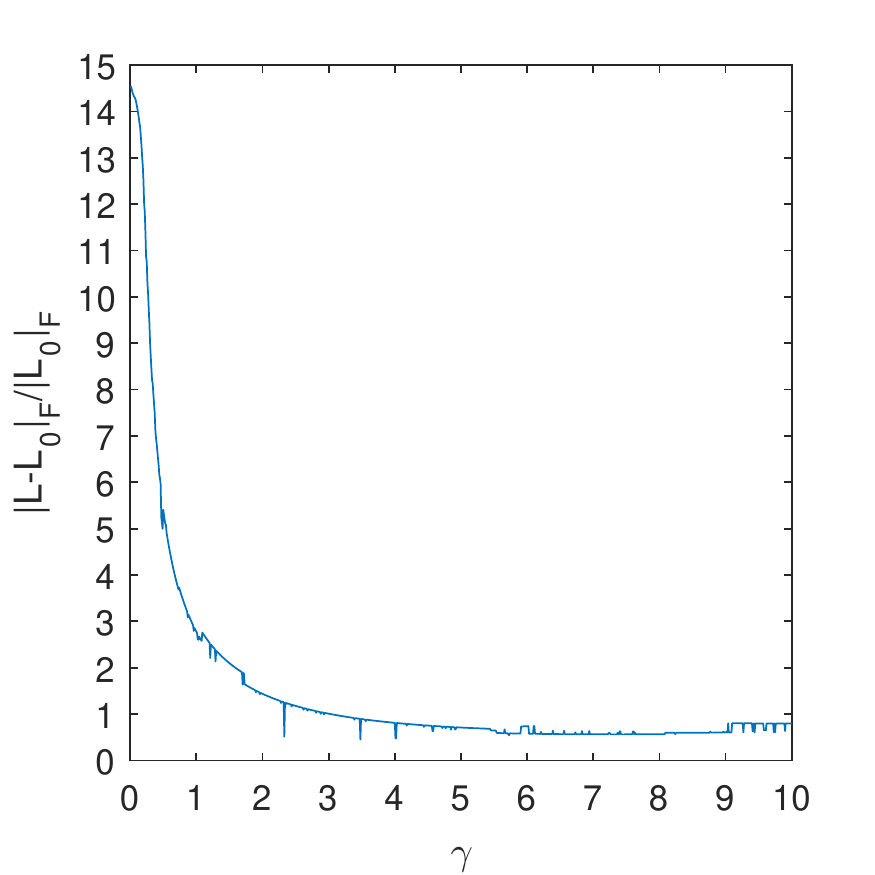}
	\end{center}
	\vspace{-.6cm}
	\caption{Relative error ($\frac{\|\mathbf{L}-\mathbf{L}_0\|_F}{\|\mathbf{L}_0\|_F}$) of FRPCAG for $\gamma\in[0.01,10]$.}
	\label{fig:frpcagParam}
	\vspace{-.3cm}
\end{figure}

For RPCAG and FRPCAG, the graphs are built using $k$-nearest neighbors. Using Euclidean distances, each sample is connected to 10 nearest neighbors with weight $e^{-\frac{s^2}{\sigma^2}}$, where $s$ is the Euclidean distance between the two samples and $\sigma$ is the average of $s$. Weight between unconnected samples is set to 0. Having obtained such weight matrix $\mathbf{A}$, we can calculate the normalised graph Laplacian $\mathbf{\Phi}=\mathbf{I}-\mathbf{D}^{-\frac{1}{2}}\mathbf{A}\mathbf{D}^{-\frac{1}{2}}$, where $\mathbf{D}$ is the diagonal degree matrix. The tolerance threshold for RPCAG and FRPCAG are all set to $\epsilon = 10^{-7}$ for reasons of consistency. We choose $\lambda=1/\sqrt{\max(n_1,n_2)}$ for a general matrix of dimension $n_1\times n_2$ as suggested in \cite{Shahdi15,Shahdi16}. For simulation experiments, $\gamma$ in RPCAG is given by the minimiser (at $\gamma=0.2$) of $\frac{\|\mathbf{L}-\mathbf{L}_0\|_F}{\|\mathbf{L}_0\|_F}$ on the benchmark problem (Figure \ref{fig:rpcagParam}). And for real-world datasets, $\gamma$ is set to 10 following \cite{Shahdi15}. For FRPCAG, we take $\gamma=\gamma_1=\gamma_2$ which is searched over $[0.01,10]$ on the benchmark problem (Figure \ref{fig:frpcagParam}). The resulting minimiser (at $\gamma=7.3$) of $\frac{\|\mathbf{L}-\mathbf{L}_0\|_F}{\|\mathbf{L}_0\|_F}$ is used in both simulation and real-world experiments.

\clearpage

\begin{figure}[h]
	\begin{center}
		\includegraphics[width=1\linewidth]{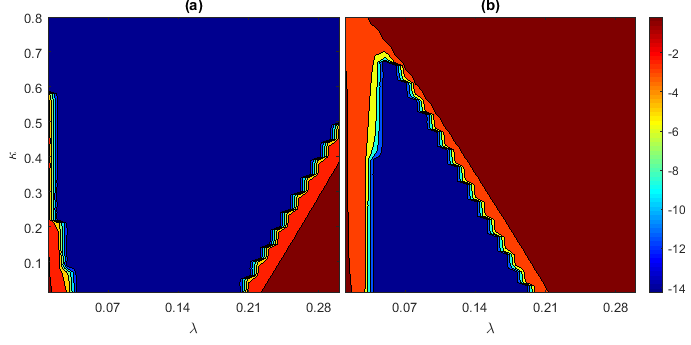}
	\end{center}
	\caption{Relative error ($\frac{\Vert\mathbf{L}-\mathbf{L}_0\Vert_F}{\Vert\mathbf{L}_0\Vert_F}$) of PCPS: \textbf{(a)} when side information is perfect; \textbf{(b)} when side information is the observation.}
	\label{fig:pcpsParam}
\end{figure}

To find $\lambda$ and $\kappa$ in PCPS, a parameter sweep in the $\kappa-\lambda$ space for perfect side information ($\mathbf{W}=\mathbf{L}_0$) is shown in Figure \ref{fig:pcpsParam} (a) and for observation as side information ($\mathbf{W}=\mathbf{M}$) in Figure \ref{fig:pcpsParam} (b) to impart a lower bound and a upper bound respectively. It can be easily seen that $\lambda=1/\sqrt{200}$ from PCP works well in both cases. Conversely, $\kappa$ depends on the quality of the side information. At $\lambda=1/\sqrt{200}$, the minimiser of $\frac{\|\mathbf{L}-\mathbf{L}_0\|_F}{\|\mathbf{L}_0\|_F}$ occurs at $\kappa=0.2$ for noisy side information. This value of $\kappa$ together with $\lambda=1/\sqrt{200}$ is used in simulation experiments for both PCPS and PCPSF. For public video sequences, increasing the value of $\kappa$ to $0.5$ can produce visual results that are noticeable to the naked eye.

\clearpage

\section{Simulation Results}

\begin{figure}[h]
	\vspace*{-0.4cm}
	\begin{center}
		\includegraphics[width=1\linewidth]{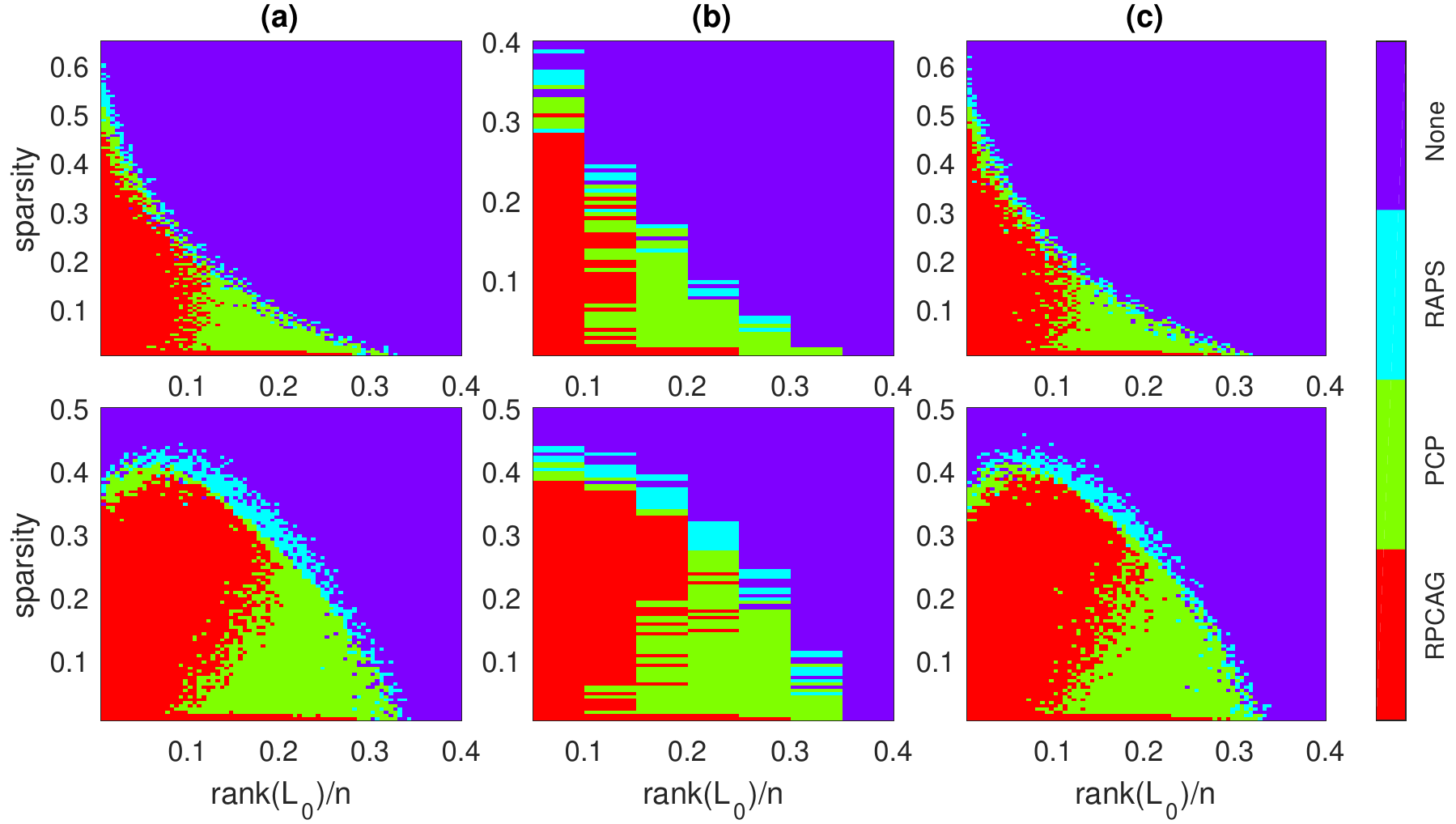}
	\end{center}
	\vspace*{-0.6cm}
	\caption{Domains of recovery by various algorithms: random signs in row \textbf{I} and coherent signs in row \textbf{II}. \textbf{(a)} for entry-wise corruptions, \textbf{(b)} for deficient ranks and \textbf{(c)} for distorted singular values.}
	\label{fig:three}
\end{figure}

\begin{figure}[b!]
	\begin{center}
		\vspace*{-0.6cm}
		\includegraphics[width=1\linewidth]{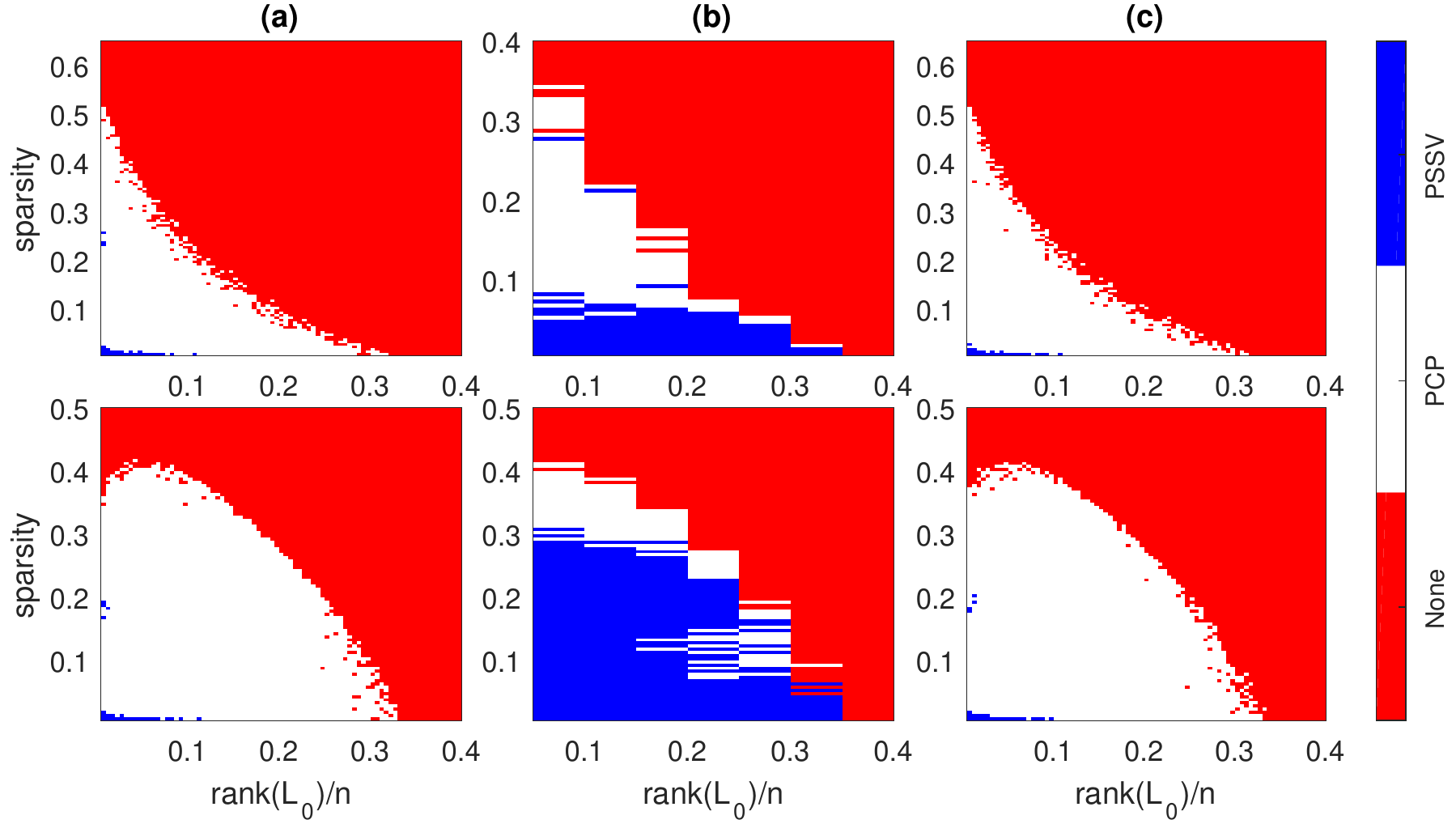}
	\end{center}
	\vspace*{-0.6cm}
	\caption{Domains of recovery by PSSV: random signs in row \textbf{I} and coherent signs in row \textbf{II}. \textbf{(a)} for entry-wise corruptions, \textbf{(b)} for deficient ranks and \textbf{(c)} for distorted singular values.}
	\label{fig:two}
	\vspace*{-0.4cm}
\end{figure}

A direct comparison of RAPS, RPCAG and PCP from simulation studies is presented in Figure \ref{fig:three}. Simulation results for PSSV are shown in Figure \ref{fig:two}.

\clearpage

\section{Real-world applications}

\subsection{Data sources}
The datasets used herein are listed below:\\\\
The Extended Yale Face Database B: \url{http://vision.ucsd.edu/~iskwak/ExtYaleDatabase/ExtYaleB.html}.\\\\
Performance  Evaluation  of  Tracking  and  Surveillance  Workshop 2006: \url{http://www.cvg.reading.ac.uk/PETS2006/data.html}.\\\\
I2R Dataset: \url{http://perception.i2r.a-star.edu.sg/bk\_model/bk\_index.html}.\\\\
The CMU Multi-PIE Face Database: \url{http://www.cs.cmu.edu/afs/cs/project/PIE/MultiPie/Multi-Pie/Home.html}.\\\\
The Extended Cohn-Kande Dataset (CK+): \url{http://www.consortium.ri.cmu.edu/ckagree/}.

\subsection{Face denoising}

\begin{figure}[h]
	\begin{center}
		\includegraphics[width=1\linewidth]{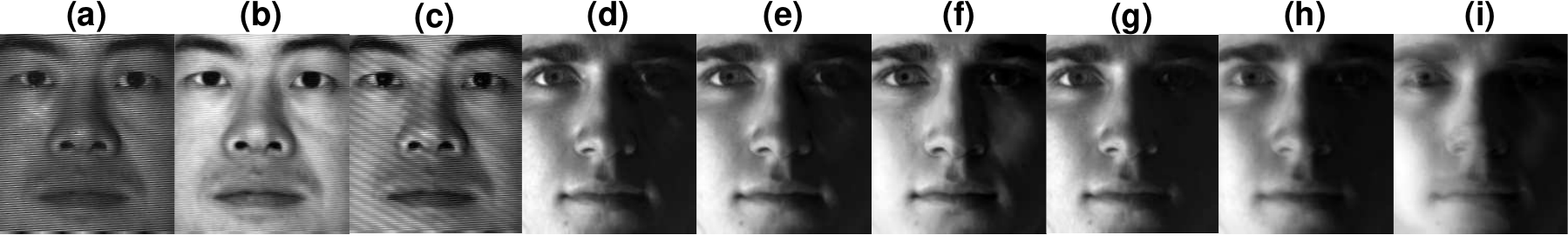}
	\end{center}
	\vspace*{-0.6cm}
	\caption{Comparison of face denoising ability: \textbf{(a,d)} single-person PSSV; \textbf{(b,e)} single-person RPCAG; \textbf{(c,f)} single-person FRPCAG; \textbf{(g)} multi-person PSSV; \textbf{(h)} multi-person RPCAG; and \textbf{(i)} multi-person FRPCAG;.}
	\label{fig:Yale2}
\end{figure}

Illustration of face denoising ability of PSSV, RPCAG, FRPCAG is presented in Figure \ref{fig:Yale2}. The average running times of different algorithms for a single subject and multiple subjects are summarised in Table \ref{table:runtime} \footnote{All experiments were performed on a 3.60GHz quad-core computer with 16GB RAM running MATLAB R2016a.}. 
\begin{table}[h]
	\begin{center}
		\begin{tabular}{|c|c|c|}
			\hline
			\multirow{ 2}{*}{Algorithm} &  \multicolumn{2}{|c|}{Time} \\
			\cline{2-3}
			& Single Subject & Multiple Subjects\\
			\hline
			K-SVD (X) & 9 min & --- \\
			\hline
			K-SVD (Y) & 78 min & --- \\
			\hline
			PCP & 12s & 5 min\\
			\hline
			PCPS & 27s & 12 min\\
			\hline
			PCPF & 16s & 9 min\\
			\hline
			PCPSF & 19s & 8 min\\
			\hline
			PSSV & 13s & 5 min\\
			\hline
			k-NN (X) & 7s & 4 min\\
			\hline
			k-NN (Y) & 1s & 8s\\
			\hline
			RPCAG & 2min & 17 min\\
			\hline
			FRPCAG & 8s & 1 min\\
			\hline
		\end{tabular}
		\caption{Running times of various algorithms.}
		\label{table:runtime}
	\end{center}
\end{table}

\clearpage

\subsection{Background Subtraction}

\begin{figure}[h]
	\begin{center}
		\includegraphics[width=1\linewidth]{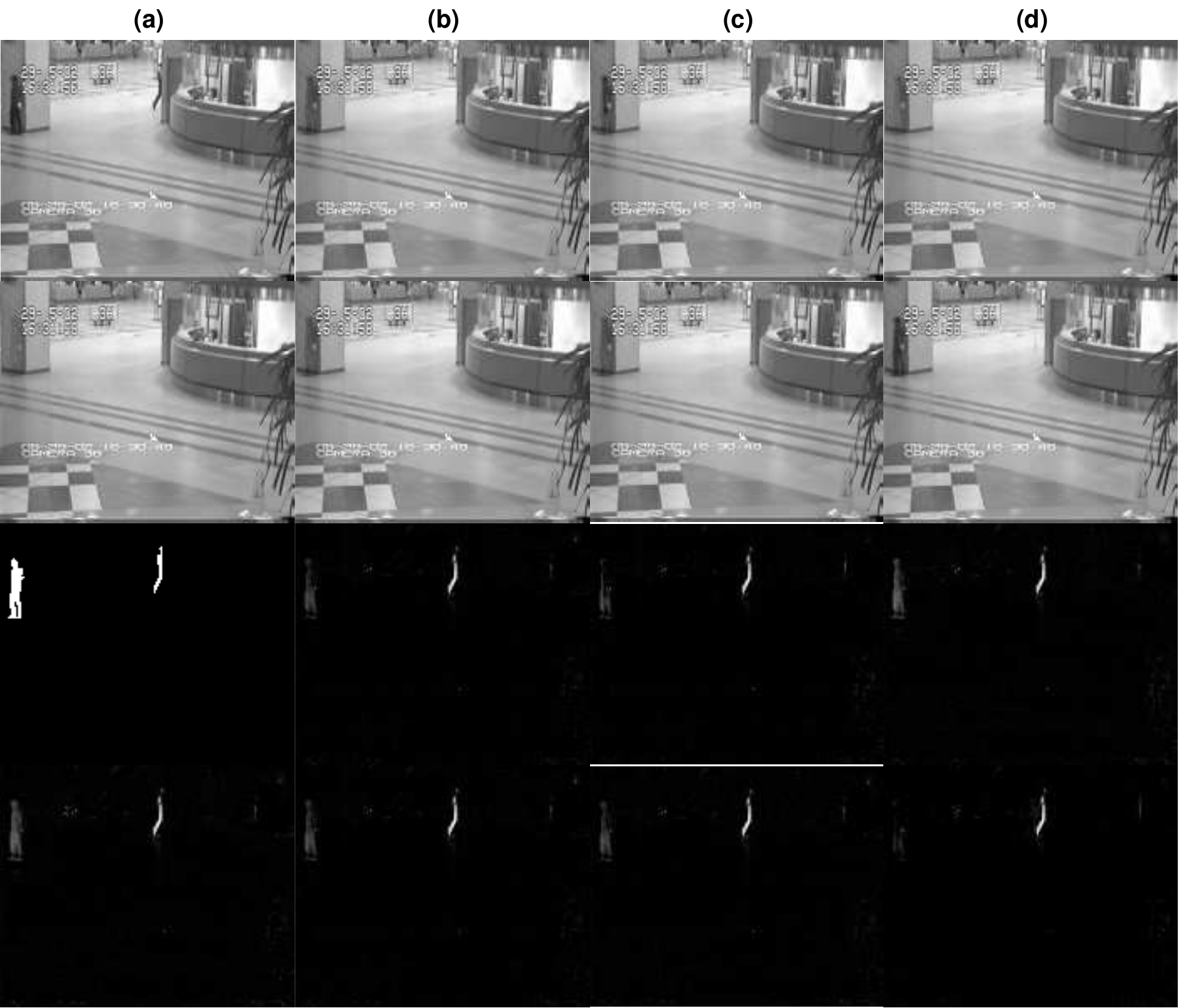}
	\end{center}
	\vspace*{-0.6cm}
	\caption{Background subtraction results for Airport : row \textbf{I} \textbf{(a)} original image; row \textbf{III} \textbf{(a)} ground truth; row \textbf{I,III} \textbf{(b)} PCP; row \textbf{I,III} \textbf{(c)} PCP (60 frames); \textbf{I,III} \textbf{(d)} PCPS (60 frames); row \textbf{II,IV} \textbf{(a)} PCPS; row \textbf{II,IV} \textbf{(b)} PSSV; row \textbf{II,IV} \textbf{(c)} RPCAG; row \textbf{II,IV} \textbf{(d)} FRPCAG.}
	\label{fig:bga}
\end{figure}

\begin{figure}[h]
	\begin{center}
		\includegraphics[width=1\linewidth]{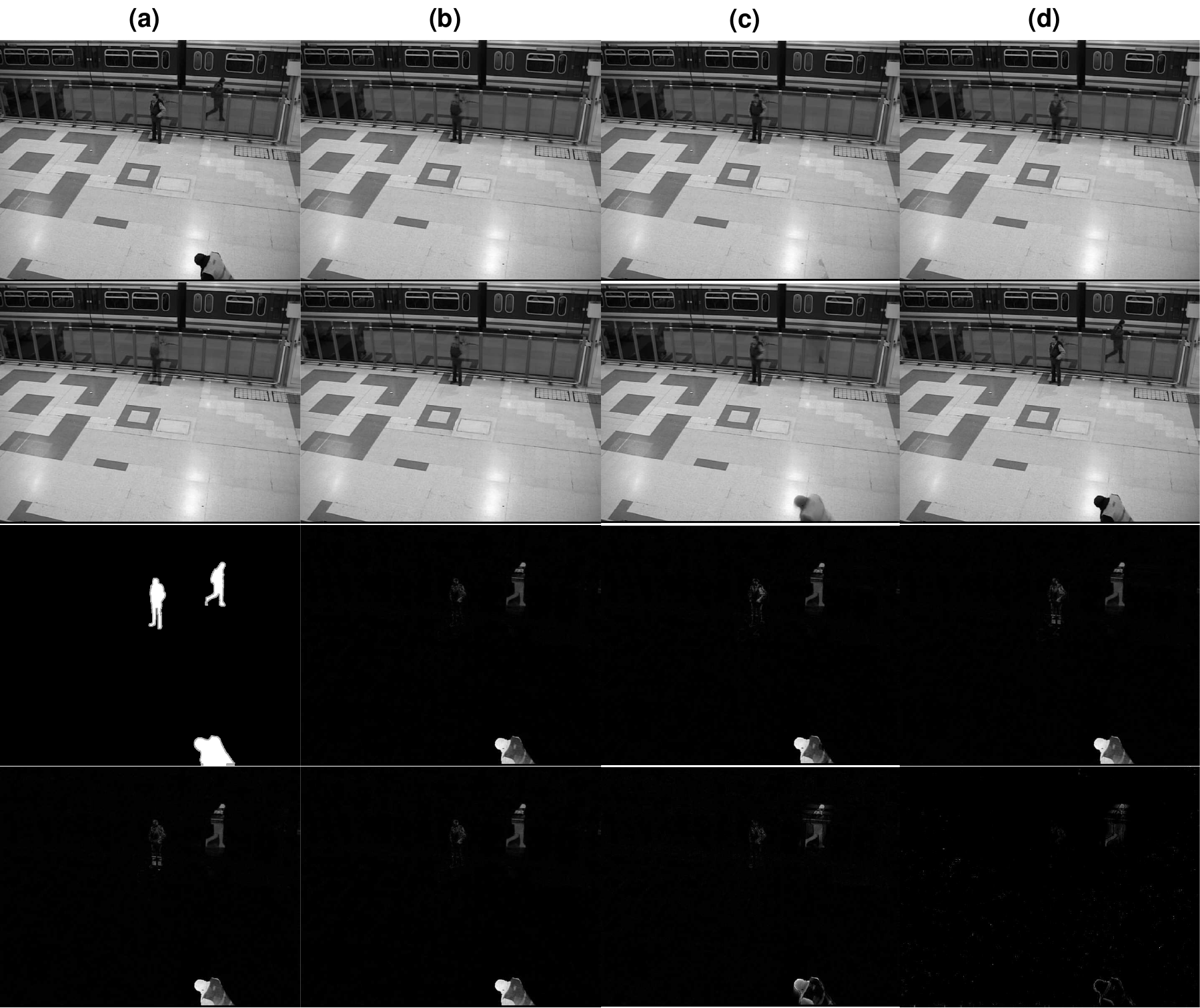}
	\end{center}
	\vspace*{-0.6cm}
	\caption{Background subtraction results for PETS : row \textbf{I} \textbf{(a)} original image; row \textbf{III} \textbf{(a)} ground truth; row \textbf{I,III} \textbf{(b)} PCP; row \textbf{I,III} \textbf{(c)} PCP (60 frames); \textbf{I,III} \textbf{(d)} PCPS (60 frames); row \textbf{II,IV} \textbf{(a)} PCPS; row \textbf{II,IV} \textbf{(b)} PSSV; row \textbf{II,IV} \textbf{(c)} RPCAG; row \textbf{II,IV} \textbf{(d)} FRPCAG.}
	\label{fig:bgb}
	\vspace*{-0.3cm}
\end{figure}

Recovered images of the background and the foreground from all methods are listed in Figure \ref{fig:bga} for Airport and Figure \ref{fig:bgb} for PETS. The running times of different algorithms for Airport and PETS are summarised in Table \ref{table:runtime2}.

\begin{table}[h]
	\begin{center}
		\begin{tabular}{|c|c|c|}
			\hline
			\multirow{ 2}{*}{Algorithm} &  \multicolumn{2}{|c|}{Time} \\
			\cline{2-3}
			& Airport & PETS\\
			\hline
			PCP & 52s & 17 min\\
			\hline
			PCPS & 2 min & 36 min\\
			\hline
			PSSV & 51s & 17 min\\
			\hline
			k-NN (X) & 52s & 2h\\
			\hline
			k-NN (Y) & 1s & 24s\\
			\hline
			RPCAG & 7 min & 3h\\
			\hline
			FRPCAG & 11s & 34s\\
			\hline
			PCP (60 frames) & 52s & 3 min\\
			\hline
			PCPS (60 frames) & 20s & 7 min\\
			\hline
		\end{tabular}
		\caption{Running times of various algorithms.}
		\label{table:runtime2}
	\end{center}
\end{table}

\clearpage

\section{Derivations}

Here we give deviations of the various equivalent subproblems for the algorithm quoted in the text:

\begin{equation*}
	\begin{split}
		&\argmin_{H}l(H,E,S,Z,N)\\=&\argmin_{H}\ ||H||_*+\kappa||E||_*+\lambda||S||_1+\langle Z,M-S-XHY^T\rangle +\frac{\mu}{2}||M-S-XHY^T||_F^2\\
		&+\langle N,H - E -X^TWY\rangle+\frac{\mu}{2}||H - E -X^TWY||_F^2\\
		=&\argmin_{H}\ ||H||_*+\langle Z,M-S-XHY^T\rangle +\frac{\mu}{2}||M-S-XHY^T||_F^2\\
		&+\langle N,H - E -X^TWY\rangle+\frac{\mu}{2}||H - E -X^TWY||_F^2\\
		=&\argmin_{H}\ ||H||_*+\tr(Z^T(M-S-XHY^T))\\ &+\frac{\mu}{2}\tr((M-S-XHY^T)^T(M-S-XHY^T))+\tr(N^T(H - E -X^TWY))\\
		&+\frac{\mu}{2}\tr((H - E -X^TWY)^T(H - E -X^TWY))\\
		=&\argmin_{H}\ ||H||_* - \tr(Z^TXHY^T) + \tr(N^TH)\\ &+\frac{\mu}{2}\tr(YH^TX^TXHY^T-YH^TX^T(M-S)-(M-S)^TXHY^T)\\
		&+\frac{\mu}{2}\tr((H - E -X^TWY)^TX^TX(H - E -X^TWY)Y^TY)\\
		=&\argmin_{H}\ ||H||_*+\mu\tr(-\frac{1}{\mu}Z^TXHY^T) + \mu\tr(\frac{1}{\mu}N^TX^TXHY^TY)\\ &+\frac{\mu}{2}\tr(YH^TX^TXHY^T-YH^TX^T(M-S)-(M-S)^TXHY^T)\\
		&+\frac{\mu}{2}\tr(YH^TX^TXHY^T-YH^TX^TX(E+X^TWY)Y^T\\
		&- Y(E+X^TWY)^TX^TXHY^T)\\
		=&\argmin_{H}\ ||H||_*+\mu\tr(YH^TX^TXHY^T-\frac{1}{2}YH^TX^T(M-S)-\frac{1}{2}(M-S)^TXHY^T\\
		&-\frac{1}{2}YH^TX^TX(E+X^TWY)Y^T-\frac{1}{2}Y(E+X^TWY)^TX^TXHY^T\\
		&-\frac{1}{2\mu}YH^TX^TZ-\frac{1}{2\mu}Z^TXHY^T+\frac{1}{2\mu}YH^TX^TXNY^T+\frac{1}{2\mu}YN^TX^TXHY^T)\\
		=&\argmin_{H}\ ||H||_*+\mu\tr((\frac{1}{2}(M-S+XEY^T+W+\frac{1}{\mu}(Z-XNY^T))-XHY^T)^T\\
		&(\frac{1}{2}(M-S+XEY^T+W+\frac{1}{\mu}(Z-XNY^T))-XHY^T))\\
		=&\argmin_{H}\ ||H||_*+\mu||\frac{1}{2}(M-S+W+\frac{1}{\mu}Z+X(E-\frac{1}{\mu}N)Y^T)-XHY^T||_F^2\\
	\end{split}
\end{equation*}

\begin{equation*}
	\begin{split}
		&\argmin_{E}l(H,E,S,Z,N)\\=&\argmin_{E}\ ||H||_*+\kappa||E||_*+\lambda||S||_1+\langle Z,M-S-XHY^T\rangle +\frac{\mu}{2}||M-S-XHY^T||_F^2\\
		&+\langle N,H - E -X^TWY\rangle+\frac{\mu}{2}||H - E -X^TWY||_F^2\\
		=&\argmin_{E}\ \kappa||E||_*+\langle N,H - E -X^TWY\rangle+\frac{\mu}{2}||H - E -X^TWY||_F^2\\
		=&\argmin_{E}\ \kappa||E||_*+\tr(N^T(H - E -X^TWY))\\
		&+\frac{\mu}{2}\tr((H - E -X^TWY)^T(H - E -X^TWY))\\
		=&\argmin_{E}\ \kappa||E||_*+\frac{\mu}{2}\tr(-\frac{2}{\mu}N^TE)\\
		&+\frac{\mu}{2}\tr(E^TE -E^T(H-X^TWY) - (H-X^TWY)^TE)\\
		=&\argmin_{E}\ \kappa||E||_*\\
		&+\frac{\mu}{2}\tr(E^TE -E^T(H-X^TWY) - (H-X^TWY)^TE-\frac{1}{\mu}E^TN-\frac{1}{\mu}N^TE)\\
		=&\argmin_{E}\ \kappa||E||_*+\frac{\mu}{2}\tr((H-X^TWY+\frac{1}{\mu}N -E)^T(H-X^TWY+\frac{1}{\mu}N -E))\\
		=&\argmin_{E}\ \kappa||E||_*+\frac{\mu}{2}||H-X^TWY+\frac{1}{\mu}N -E||_F^2\\
	\end{split}
\end{equation*}

\begin{equation*}
	\begin{split}
		&\argmin_{S}l(H,E,S,Z,N)\\=&\argmin_{S}\ ||H||_*+\kappa||E||_*+\lambda||S||_1+\langle Z,M-S-XHY^T\rangle +\frac{\mu}{2}||M-S-XHY^T||_F^2\\
		&+\langle N,H - E -X^TWY\rangle+\frac{\mu}{2}||H - E -X^TWY||_F^2\\
		=&\argmin_{S}\ \lambda||S||_1+\langle Z,M-S-XHY^T\rangle+\frac{\mu}{2}||M-S-XHY^T||_F^2\\
		=&\argmin_{S}\ \lambda||S||_1+\tr( Z^T(M-S-XHY^T))\\
		&+\frac{\mu}{2}\tr((M-S-XHY^T)^T(M-S-XHY^T))\\
		=&\argmin_{S}\ \lambda||S||_1+\frac{\mu}{2}\tr(-\frac{2}{\mu} Z^TS)\\
		&+\frac{\mu}{2}\tr(S^TS-S^T(M-XHY^T) - (M-XHY^T)^TS)\\
		=&\argmin_{S}\ \lambda||S||_1\\
		&+\frac{\mu}{2}\tr(S^TS-S^T(M-XHY^T) - (M-XHY^T)^TS-\frac{1}{\mu} S^TZ-\frac{1}{\mu} Z^TS)\\
		=&\argmin_{S}\ \lambda||S||_1+\frac{\mu}{2}\tr((M-XHY^T+\frac{1}{\mu}Z-S)^T(M-XHY^T+\frac{1}{\mu}Z-S))\\
		=&\argmin_{S}\ \lambda||S||_1+\frac{\mu}{2}||M-XHY^T+\frac{1}{\mu}Z-S||_F^2\\
	\end{split}
\end{equation*}

\clearpage

\section{Further comments}

One might suggest that a potentially better and more direct approach in using the side information is to subtract the side information. That is, do RPCA on $\mathbf{M}'=\mathbf{M}-\mathbf{W}$, where $\mathbf{M}$ is the data and $\mathbf{W}$ is the noisy side information, to obtain $\mathbf{M}'=\mathbf{L}'+\mathbf{S}$ with $\mathbf{L}=\mathbf{L}'+\mathbf{W}$.

We argue that this is not correct for the following reasons:
\begin{itemize}
	\item The rank of $\mathbf{L}'$ is no smaller than $\mathbf{L}$, which does not make the problem any simpler than the original one.
	\item When $\mathbf{W}$ is merged into $\mathbf{M}$, the additional information provided by $\mathbf{W}$ is lost and the features can on longer be applied.
	\item When $\mathbf{W}$ includes full-rank noise on $\mathbf{L}$, $\mathbf{L}'$ is not low-rank anymore. This violates the assumption of RPCA.
\end{itemize}

\begin{figure}[h!]
	\vspace*{-1.6cm}
	\begin{center}
		\includegraphics[width=1\linewidth]{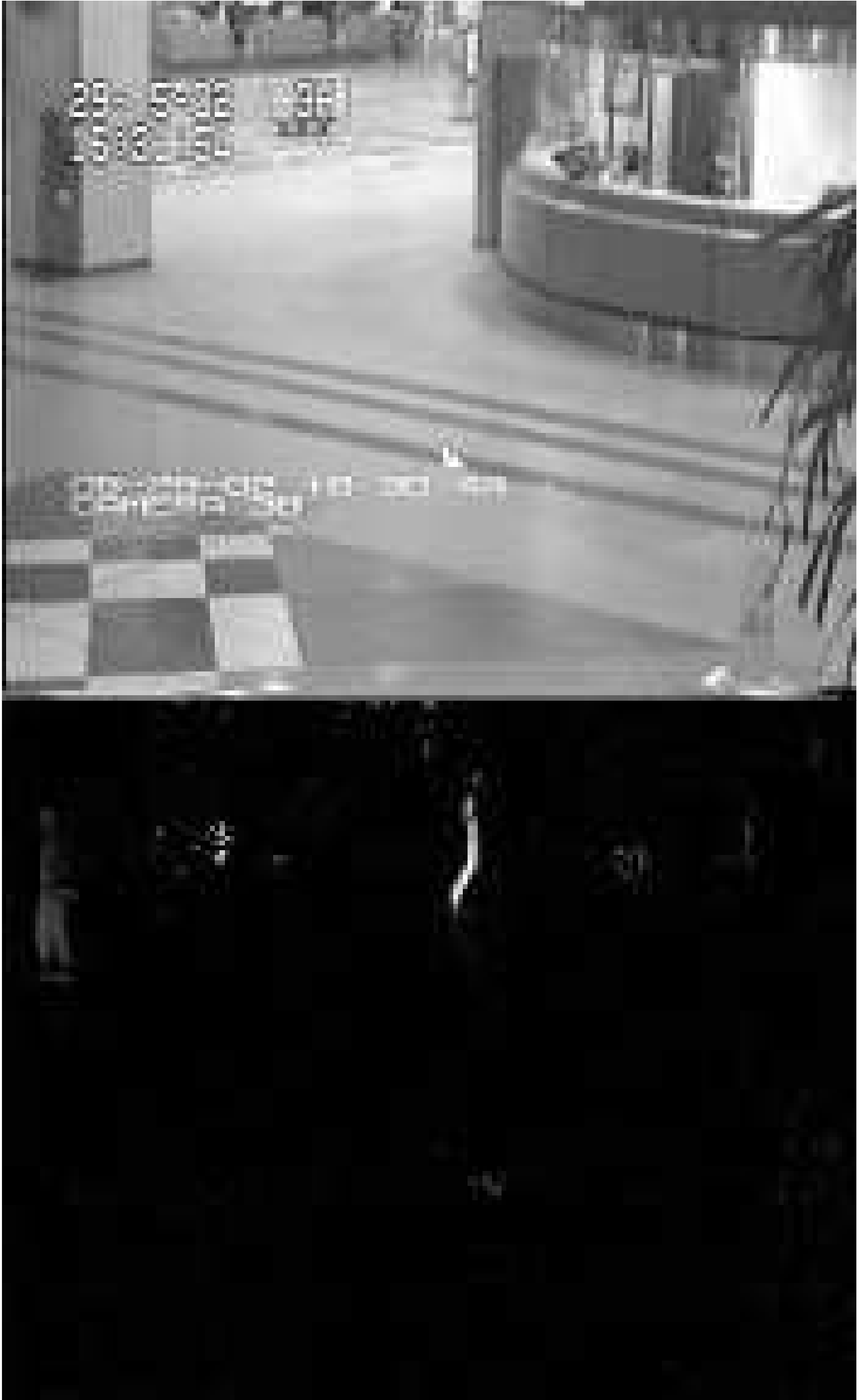}
	\end{center}
	\vspace*{-0.6cm}
	\caption{Background subtraction by suggestion: background in row $\mathbf{I}$ and segmentaion in row $\mathbf{II}$.}
	\label{fig:artificial1}
\end{figure}

\begin{figure}[h!]
	\vspace*{-1.6cm}
	\begin{center}
		\includegraphics[width=1\linewidth]{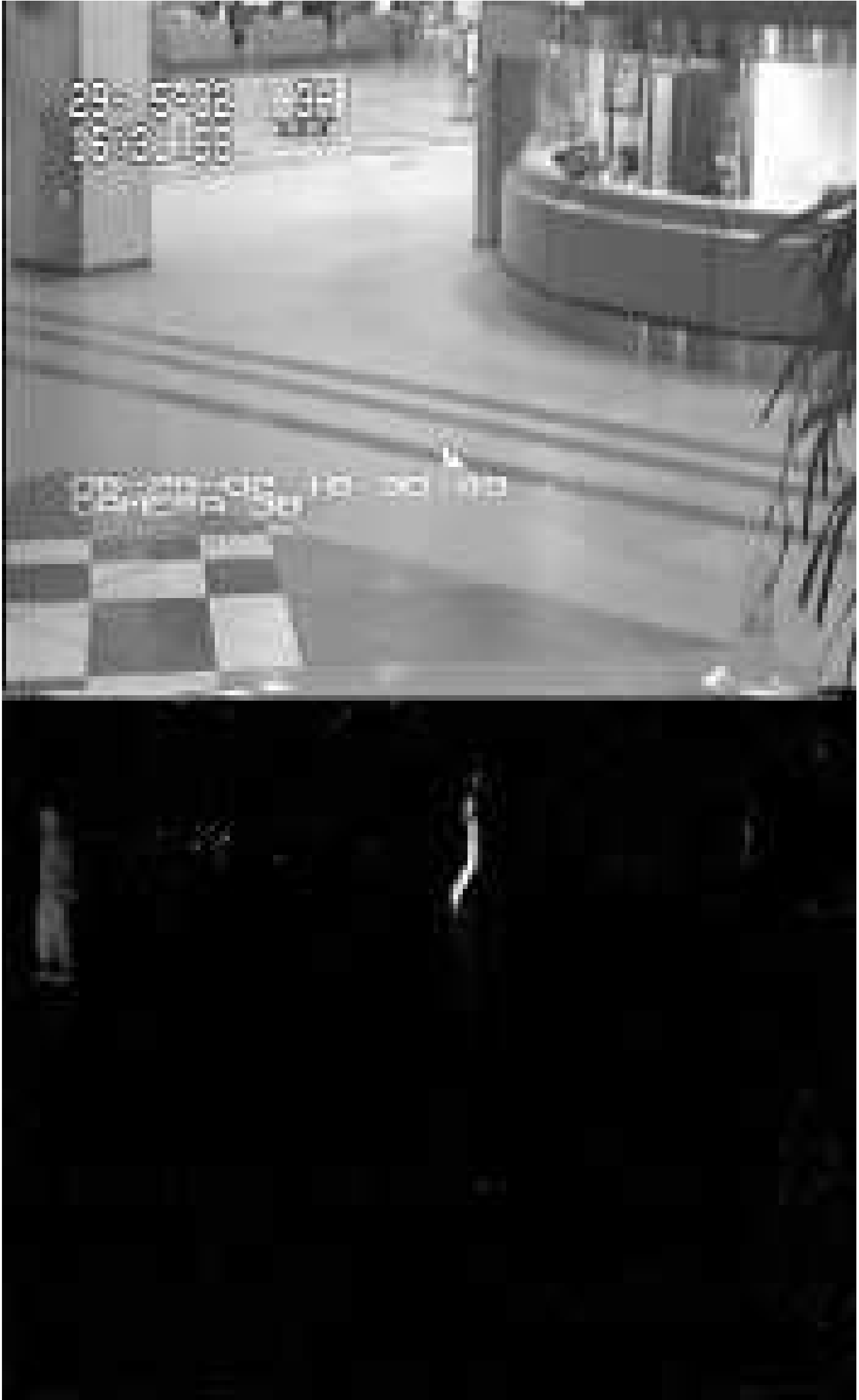}
	\end{center}
	\vspace*{-0.6cm}
	\caption{Background subtraction by PCPS: background in row $\mathbf{I}$ and segmentaion in row $\mathbf{II}$.}
	\label{fig:artificial2}
\end{figure}

\begin{figure}[h!]
	\vspace*{-1.6cm}
	\begin{center}
		\includegraphics[width=1\linewidth]{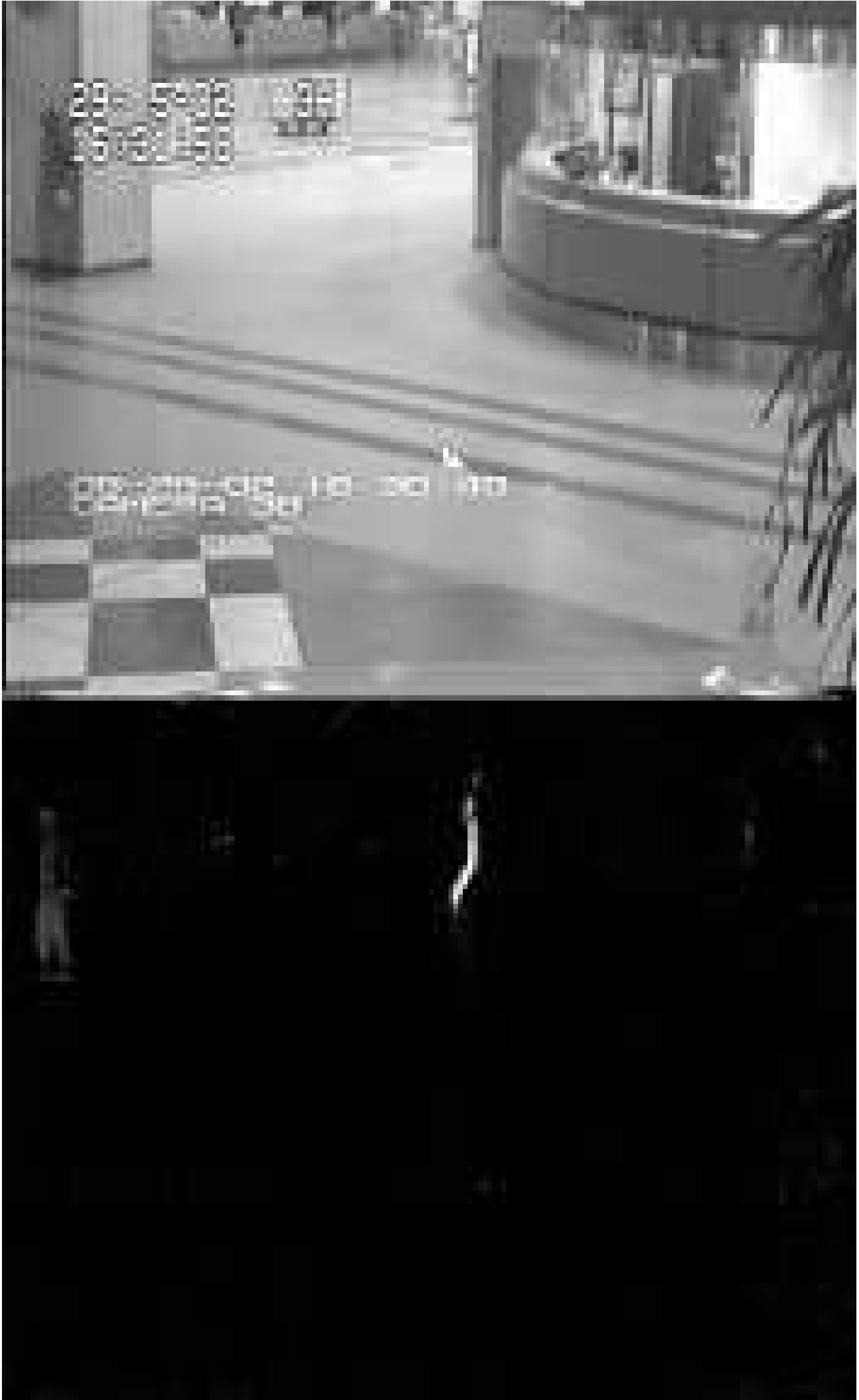}
	\end{center}
	\vspace*{-0.6cm}
	\caption{Background subtraction by PCP: background in row $\mathbf{I}$ and segmentaion in row $\mathbf{II}$.}
	\label{fig:artificial3}
\end{figure}

To verify our claim, we perform the Airport experiment again, but with different side information than that used in the paper. We collect 200 different frames of relatively clean backgrounds and stack them into the side information $\mathbf{W}$. Comparison of the suggestion with PCPS and PCP is shown in Figure \ref{fig:artificial1}, \ref{fig:artificial2} and \ref{fig:artificial3}. It is clearly visible that the low-rank structure cannot be recovered by the suggestion and spurious noises are introduced in the segmentation, whereas PCPS works impeccably segmenting accurately the foreground moving objects leaving a clean background.

\clearpage

{\small
\bibliographystyle{ieee}
\bibliography{egbib}
}

\end{document}